\documentclass{article}       

\usepackage{arxiv}

\usepackage{times}
\usepackage{xcolor}
\usepackage{graphicx}
\usepackage{subcaption}
\usepackage{array} 
\usepackage{tabularx}
\usepackage{tabu}
\usepackage{bbm}
\usepackage{amsfonts}
\usepackage{nicefrac}       
\usepackage{microtype}      
\usepackage{amsthm}
\usepackage{appendix}
\usepackage{mathtools}
\usepackage{url}
\usepackage{tikz}
\usepackage{smartdiagram}
\usepackage{hyperref}
\usetikzlibrary{arrows}
\usepackage{amssymb}
\usepackage{algorithm} 
\usepackage{algorithmic}
\usepackage[algo2e,ruled,vlined]{algorithm2e}
\usepackage{algorithm}
\usepackage{algorithmic}
\usepackage{multirow}
\usepackage{textcomp}
\usepackage{setspace}
\usepackage[utf8]{inputenc}
\usepackage[T1]{fontenc}
\usepackage{cases}
\usepackage{color}

\usepackage{empheq}
\usepackage[most]{tcolorbox}

\let\proof\relax

\newtheorem{prop}{Proposition}

\DeclareFontFamily{U}{wncy}{}
\DeclareFontShape{U}{wncy}{m}{n}{<->wncyr10}{}
\DeclareSymbolFont{mcy}{U}{wncy}{m}{n}
\DeclareMathSymbol{\Sh}{\mathord}{mcy}{"58} 

\def \imageDir{./images/}

\def \Expect{\mathbb{E}}
\def \R2{\mathbb{R}^2}
\def \R{\mathbb{R}}

\newcommand{\mySet}[1]{\mathcal{#1}}

\newcommand{\indic}[1]{\mathbbm{1}_{#1}}

\appendixtitleon
\appendixtitletocon

\newtcbox{\mymath}[1][]{%
    nobeforeafter, math upper, tcbox raise base,
    enhanced, colframe=blue!30!black,
    colback=blue!05, boxrule=1pt,
    #1}

\DeclareMathOperator*{\argmin}{\arg\!\min}

\newcolumntype{C}{>{\centering\arraybackslash}X}


\author{Alasdair Newson\\
  LTCI, T\'el\'ecom ParisTech, Universit\'e Paris Saclay\\
  46 rue Barrault, 75013, Paris\\
  \texttt{alasdair.newson@telecom-paristech.fr}\\
  \And
  Andrés Almansa$^\ast$ \\
  CNRS MAP5, Universit\'e Paris Descartes\\
  45, rue des Saints-P\`eres, 75006, Paris \\
  \texttt{andres.almansa@parisdescartes.fr}\\
  \And
  Yann Gousseau$^\ast$\\
  LTCI, T\'el\'ecom ParisTech, Universit\'e Paris Saclay\\
  46 rue Barrault, 75013, Paris\\
  \texttt{yann.gousseau@telecom-paristech.fr}\\
  \And
  Saïd Ladjal\thanks{Indicates equal contribution}\\
  LTCI, T\'el\'ecom ParisTech, Universit\'e Paris Saclay\\
  46 rue Barrault, 75013, Paris\\
  \texttt{said.ladjal@telecom-paristech.fr}\\
}

\title{Processsing Simple Geometric Attributes with Autoencoders}

\begin{document}
\maketitle

\begin{abstract}

Image synthesis is a core problem in modern deep learning, and many recent architectures such as autoencoders and Generative Adversarial networks produce spectacular results on highly complex data, such as images of faces or landscapes. While these results open up a wide range of new, advanced synthesis applications, there is also a severe lack of theoretical understanding of how these networks work. This results in a wide range of practical problems, such as difficulties in training, the tendency to sample images with little or no variability, and generalisation problems. In this paper, we propose to analyse the ability of the simplest generative network, the autoencoder, to encode and decode two simple geometric attributes : size and position. We believe that, in order to understand more complicated tasks, it is necessary to first understand how these networks process simple attributes. For the first property, we analyse the case of images of centred disks with variable radii. We explain how the autoencoder projects these images to and from a latent space of smallest possible dimension, a scalar. In particular, we describe a closed-form solution to the decoding training problem in a network without biases, and show that during training, the network indeed finds this solution. We then investigate the best regularisation approaches which yield networks that generalise well. For the second property, position, we look at the encoding and decoding of Dirac delta functions, also known as ``one-hot'' vectors. We describe a hand-crafted filter that achieves encoding perfectly, and show that the network naturally finds this filter during training. We also show experimentally that the decoding can be achieved if the dataset is sampled in an appropriate manner. We hope that the insights given here will provide better understanding of the precise mechanisms used by generative networks, and will ultimately contribute to producing more robust and generalisable networks.


\end{abstract}

\section{Introduction}
\label{sec:intro}

Image synthesis is a central issue of modern deep learning, and in particular encoder-decoder neural networks (NNs), which include many popular networks such as autoencoders, Generative Adversarial Networks (GANs) \cite{goodfellow2014generative}, variational autoencoders \cite{Kingma2014Auto} etc. These networks are able to produce truly impressive results \cite{radford2015unsupervised,Zhu2016Generative,choi2017stargan,isola2017image,karras2017progressive}. However, as in many areas of deep learning, there is a severe lack of theoretical understanding of the networks. In practice, this means that these approaches suffer from a variety of problems, such as difficulty in training the networks \cite{salimans2016improved}, the tendency to sample with little or no variety (``mode collapse'' \cite{metz2016unrolled}), and generalisation problems. There is also the extremely important question of how to interpolate in the latent space (ie how to interpolate between two visual objects via the latent space), which is still an open problem and is mostly done with linear interpolation at the moment \cite{Zhu2016Generative,upchurch2017deep}. Such questions must be answered if these types of networks can be used reliably. For this, we need to understand the inner workings of these networks.

In this paper, we propose to study how the \emph{autoencoder} (the simplest generative neural network) processes two basic image properties:
\begin{itemize}
	\item size; 
	\item position;
\end{itemize}
In order to do this, we shall analyse the manner in which the autoencoder works in the case of very simple images. We believe that, in order to understand more complicated synthesis situations, it is necessary to first understand how these networks process simple attributes. For the first property, size, we will look at grey-level images of \emph{disks} with different radii, as such images represent a very simple setting for the notion of size. Secondly, we will look at images containing Dirac delta functions (vectors where one element is non-zero, also called ``one-hot vectors''), and determine how the autoencoder can extract the position from such signals. Again, this appears the simplest way to study how the spatial position of an object is processed by such networks. Studying these mechanisms is extremely important, if the networks are to be understood. A recent work by Liu et al. \cite{liu2018intriguing} also highlighted the importance of studying how NNs work in such simple cases, and their experimental study of one such case lead them to propose the ``CoordConv'' network layer. In our work, we propose a theoretical investigation of the autoencoder in the two aforementioned cases. 

There are several advantages to such an approach. Firstly, since the class of objects we consider has an explicit parametrisation, we know the optimal compression which the autoencoder should obtain. In other words, we know the minimum size of latent space which is sufficient to correctly represent the data. Most applications of autoencoders or similar networks consider relatively high-level input objects, ranging from the MNIST handwritten digits to abstract sketches of conceptual objects (\cite{Zhu2016Generative,ha2017neural}).
Secondly, the nature of our approach fixes certain architecture characteristics of the network, such as the number of layers, leaving fewer free parameters to tune. This means that the conclusions which we obtain are more likely to be robust than in the case of more high-level applications. Finally, we can analyse the generalisation capacity of the autoencoder with great precision. Indeed, a central problem of deep learning is ensuring that the network is able to generalise outside of the observed data. We are able to study how well the autoencoder does this by removing data from the training set which correspond to a certain region of the parameters, and see whether the autoencoder is able to reconstruct data in that zone.

To summarise, we propose the following contributions in this paper :
\begin{itemize}
    \item We verify that the autoencoder can correctly learn how to encode and decode a simple shape (the disk) to and from a single scalar, where the size of the disks is the varying parameter.
    \item We investigate and explain the internal mechanisms of the autoencoder which achieve this.
    \item We analyse the best regularisation approaches which lead to better generalisation in the case where certain disk sizes are not observed in the training data.
    \item We show how the autoencoder can process the \emph{position} of an object in an image. For this, we study the simple case of a Dirac (ie a one-hot vector) as an input to the network.
\end{itemize}

One of the ultimate, long-term, goals in studying the precise properties of autoencoders in simple cases such as these is to identify architectures and regularisations which yield robust autoencoders which can generalise well in regions unobserved during training. We hope that this work can contribute to attaining this goal.



\section{Prior work}
\label{sec:prior}

The concept of autoencoders has been present for some time in the learning community (\cite{LeCun1987Learning,Bourlard1988Auto}). Autoencoders are neural networks, often convolutional neural networks, whose purpose is twofold. Firstly, to compress some input data by transforming it from the input domain to another space, known as the latent (or code) space, which is learned by the network. The second goal of the autoencoder is to take this latent representation and transform it back to the original space, such that the output is similar, with respect to some criterion, to the input. In most applications, the dimensionality $d$ of the latent space is smaller than that of the original data, so that the autoencoder is encouraged to discover useful features of the data. In practice, we obviously do not know the exact value of $d$, but we would still like to impose as much structure in the latent space as possible. This idea lead to the regularisation in the latent space of autoencoders, which comes in several flavours. The first is the sparse autoencoder (\cite{Ranzato2007Sparse}), which attempts to have as few active (non-zero) neurons as possible in the network. This can be done either by modifying the loss function to include sparsity-inducing penalisations, or by acting directly on the values of the code $z$. In the latter option, one can use rectified linear units (ReLUs) to encourage zeros in the code (\cite{Glorot2011Deep}) or simply specify a maximum number of non-zero values as in the ``k-sparse'' autoencoder (\cite{Makhzani2013K}). Another approach, taken by the variational autoencoder, is to specify the a priori distribution of the code $z$. \cite{Kingma2014Auto} use the Kullback-Leibler divergence to achieve this goal, and the authors impose a Gaussian distribution on $z$.
The ``contractive'' autoencoder (\cite{Rifai2011Contractive}) encourages the derivatives of the code with respect to the input image to be small, meaning that the representation of the image should be robust to small changes in the input.

Autoencoders can be applied to a variety of problems, such as denoising (``denoising autoencoder'') or image compression (\cite{Balle2016End}). For a good overview of autoencoders, see the book of Goodfellow et al. (\cite{Goodfellow2016Deep}). Recently, a great deal of attention has been given to the capacity of GANs (\cite{goodfellow2014generative}) and autoencoders, to generate new images. In the past couple of years, increasingly impressive results have been produced by more and more complex networks \cite{Zhu2016Generative,choi2017stargan,isola2017image,karras2017progressive} It is well-known that these networks have important limitations, such as the tendency to produce low quality images or to reproduce images from the training set because of mode collapse \cite{metz2016unrolled}. Overcoming these drawbacks requires us to understand generative networks in greater depth, and the best place to start such an investigation is with simple cases, which we now proceed to analyse.


\begin{figure*}
\includegraphics[width=\textwidth]{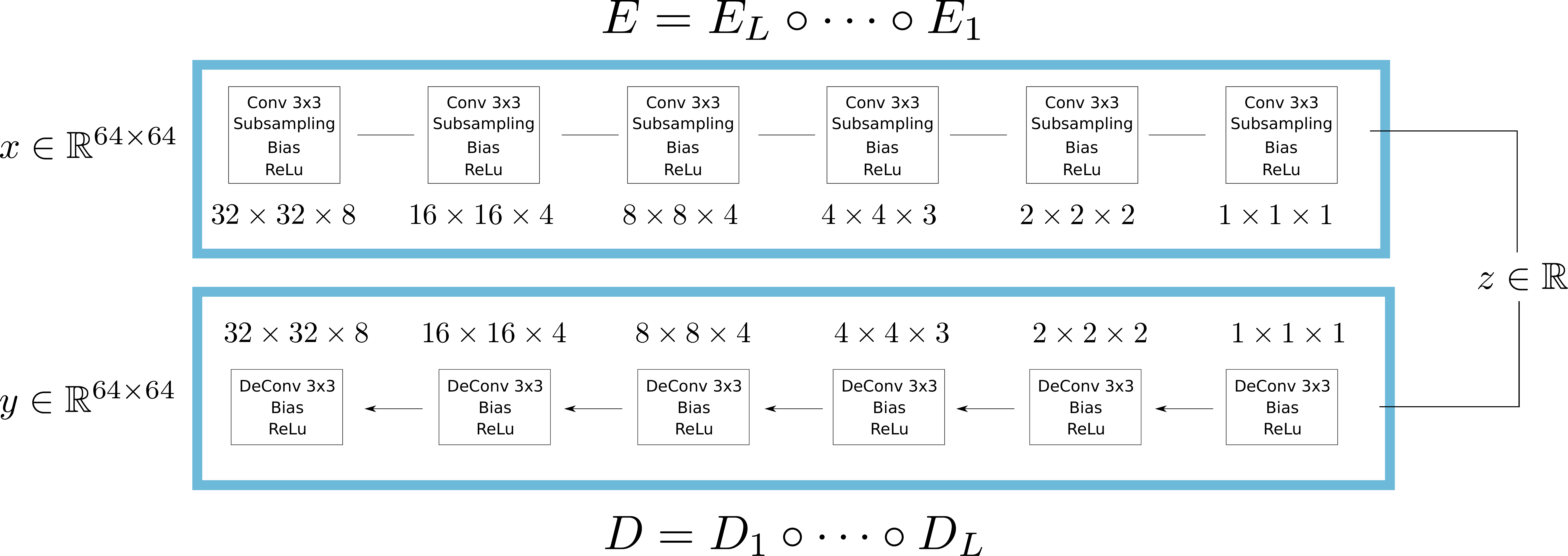}
\caption{Generic autoencoder architecture used in the geometric experiments of Section~\ref{sec:autoencodingDisks}.}
\label{fig:autoencoderArchitecture}
\end{figure*}

\begin{table}[t]
\begin{tabularx}{\linewidth}{|p{0.1\linewidth}||>{\centering\arraybackslash}C>{\centering\arraybackslash}C>{\centering\arraybackslash}
>{\centering\arraybackslash}C>{\centering\arraybackslash}C>{\centering\arraybackslash}C>{\centering\arraybackslash}C>{\centering\arraybackslash}C|}
    \hline
    Layer & Input & \multicolumn{5}{|c|}{Hidden layers} & Code ($z$)      \\
    \hline
    Depths (disk) & 1 & 8 & 4 & 4 & 3 & 2 & 1 \\
    Depths (position) & 1 & 1 & 1 & 1 & 1 & 1 & 1 \\
    \hline
\end{tabularx}
\vspace{10mm}
\begin{tabularx}{\linewidth}{|p{0.1\linewidth}||>{\centering\arraybackslash}C>{\centering\arraybackslash}C>{\centering\arraybackslash}
>{\centering\arraybackslash}C>{\centering\arraybackslash}C>{\centering\arraybackslash}C|}
	\hline
    Parameter & Spatial filter size & Non-linearity & Learning rate & Learning algorithm & Batch size \\
    \hline
    Value & $3 \times 3$ & Leaky ReLu ($\alpha=0.2$, see Eq.~\eqref{eq:lRelu}) & 0.001 & Adam & 300 \\
    \hline
\end{tabularx}
\vspace{-1cm}
\caption{Parameters of autoencoder designed for processing centred disks of random radii.}
\label{tab:parameterTable}
\end{table}

\section{Notation and Autoencoder Architecture}
\label{sec:aeArchitecture}

Although autoencoders have been extensively studied, very little is known concerning the actual inner mechanics of these networks, in other words quite simply, how they work. This is obviously much too vast a question in the general case, however very often deep learning is applied to the specific case of \emph{images}. In this work, we aim to discover how, with a cascade of simple operations common in deep networks, an autoencoder can encode and decode very simple images. In view of this goal, we propose to study in depth the case of \emph{disks} of variable radii. There are two advantages to this approach. Firstly, it allows for a full understanding in a simplified case, and secondly, the true dimensionality of the latent space is known, and therefore the architecture is constrained.

Before continuing, we describe our autoencoder in a more formal fashion.

We consider square input images, which we denote with $x \in \R^{m \times m}$, and codes $z \in \mathbb{R}^d$, and $d$ is the dimension of the latent space. The autoencoder consists of the couple $(E,D)$, the encoder and decoder which transform to and from the ``code'' space, with $E : \R^{m \times n} \rightarrow \R^{d}$ and $D : \R^{d} \rightarrow \R^{m \times n}$. As mentioned, the goal of the auto-encoder is to compress and uncompress a signal to (and from) a representation with a smaller dimensionality, while losing as little information as possible. Thus, we search for the parameters of the encoder and the decoder, which we denote with $\Theta_E$ and $\Theta_D$ respectively, by minimising
\begin{equation}
(\Theta_E, \Theta_D) =  \argmin_{\Theta_E,\Theta_D}\sum_{x} || x - D(E(x)) ||^2_2\\
\end{equation}

The autoencoder consists of a series of convolutions with filters of small compact support, sub-sampling/up-sampling, biases and non-linearities. The values of the filters are termed the weights of the network, and we denote the encoding filters with $w_{\ell,i}$, where $\ell$ is the layer and $i$ the index of the filter. Similarly, we denote the decoding filters $w'_{\ell,i}$. Since we use \emph{strided convolutions}, the subsampling is carried out just after the convolution. The encoding and decoding biases are denoted with $b_{\ell,i}$ and $b'_{\ell,i}$, and we choose leaky ReLUs for the non-linearities :
\begin{equation}
\phi_\alpha(x) = \begin{cases}
        x, & \text{for }  x \geq 0\\
        \alpha x, & \text{for } x < 0
        \end{cases},
\label{eq:lRelu}
\end{equation}
with parameter $\alpha=0.2$. Thus, the output of a given encoding layer is given by 
\begin{equation}
E^{l+1}_i = \phi_\alpha( E^{l} \ast w_{\ell,i} + b_{\ell,i} ),
\end{equation}
and similarly for the decoding layers (except for zero-padding upsampling prior to the convolution), with weights and biases $w'$ and $b'$, respectively. We have used an abuse of notation by not indicating the subsampling here, as this is carried out with the strided convolution.

We consider that the spatial support of the image $\Omega=[0,m-1] \times [0,m-1]$ is fixed throughout this work with $m=64$, and also that the subsampling rate $s$ is fixed to 2. In the encoder, subsampling is carried out until $z$ achieves the size defined by the problem at hand. In the case of disks with varying radii, it is reasonable to assume that $z$ will be a scalar. Thus, the number of layers in our encoder and decoder is not a free parameter. We set the support of all the convolutional filters in our network to $3 \times 3$. The architecture of our autoencoder remains the same throughout the paper, and is shown in Figure~\ref{fig:autoencoderArchitecture}. We summarise our parameters in Table~\ref{tab:parameterTable}.

\section{Autoencoding disks}
\label{sec:autoencodingDisks}

\def \disk_width{0.19}
\begin{figure*}
\begin{tabularx}{\textwidth}{X X X X X}
\multicolumn{5}{c}{Input} \\
\includegraphics[width=\disk_width \textwidth]{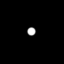}&
\includegraphics[width=\disk_width \textwidth]{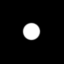}&
\includegraphics[width=\disk_width \textwidth]{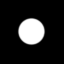}&
\includegraphics[width=\disk_width \textwidth]{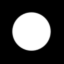}&
\includegraphics[width=\disk_width \textwidth]{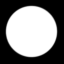}
\\
\includegraphics[width=\disk_width \textwidth]{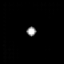}&
\includegraphics[width=\disk_width \textwidth]{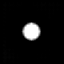}&
\includegraphics[width=\disk_width \textwidth]{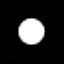}&
\includegraphics[width=\disk_width \textwidth]{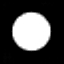}&
\includegraphics[width=\disk_width \textwidth]{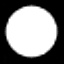}
\\
\multicolumn{5}{c}{Output}
\end{tabularx}
\caption{\textbf{Result of autoencoding disks, with a latent space dimension of size $d=1$}}
\label{fig:disk_autoencoding}
\end{figure*}

\subsection{Training dataset and preliminary autoencoder results}
\label{subsec:trainingDataDisk}

Our training set consists of grey-level images of centred disks. The radii of the disks are sampled following a uniform distribution $\mathcal{U}((0,\frac{m}{2}))$. We generate 3000 disks in the training set, so that the radius distribution is quite densely sampled. In order to create a continuous dataset, we slightly blur the disks with a Gaussian filter $g_\sigma$. The exact manner in which this is done, using a Monte Carlo simulation, is explained in Appendix\ref{app:diskDataSet}.

Theoretically, an optimal encoder would only need one scalar to represent the image. Therefore the architecture in Figure~\ref{fig:autoencoderArchitecture} is set up to ensure a code size $d=1$. 

After training, we first make two important experimental observations :
\begin{itemize}
    \item The network learns to encode/decode correctly disks with a latent space size of $d=1$;
    \item The code $z$ which is learned can be interpolated and the corresponding decoding is meaningful;
\end{itemize}
These two observations can be verified in Figure~\ref{fig:disk_autoencoding}. We now proceed to see how the autoencoder actually works on a detailed level, starting with the encoding step.


\subsection{Encoding a disk}
Encoding a centred disk of a certain radius to a scalar $z$ can be done in several ways, the most intuitive being integrating over the \emph{area} of the disk (encoding a scalar proportionate to its area) or integrating over the \emph{perimeter} of the disk (encoding a scalar proportionate to its radius). The empirical evidence given by our experiments points towards the first option, since $z$ seems to represent the area and not the radius of the input disks (see Figure~\ref{fig:disk_interpolation}). If this is the case, the integration operation can be done by means of a simple cascade of linear filters. As such, we should be able to encode the disks with a network containg only convolutions and sub-sampling, and having non-linearities. We have verified experimentally this with such an encoder. 

\def \disk_width{0.15}
\begin{figure*}
\begin{tabular}{c c c p{0.25\textwidth}}
\includegraphics[width=\disk_width \textwidth]{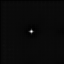}&
\includegraphics[width=\disk_width \textwidth]{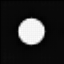}&
\includegraphics[width=\disk_width \textwidth]{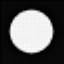}&
\multirow{2}{*}[2cm]{\includegraphics[width= 0.5\textwidth]{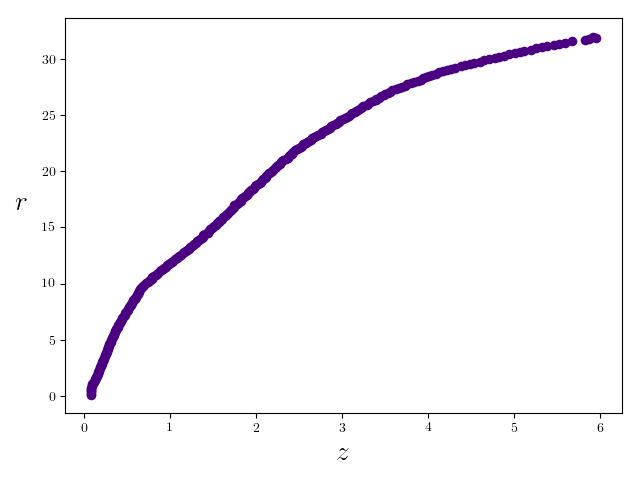}}\\
\includegraphics[width=\disk_width \textwidth]{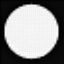}&
\includegraphics[width=\disk_width \textwidth]{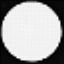}&
\includegraphics[width=\disk_width \textwidth]{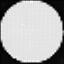}&
\\
\multicolumn{3}{c}{Decoded disks with linearly increasing $z$} & 
\end{tabular}
\caption{\textbf{Investigating the latent space in the case of disks}. On the left side, we have interpolated $z$ in the latent space between two encoded input disks (one small and one large), and show the decoded, output image. It can be seen that the training works well, with the resulting code space being meaningful. On the right, we plot the radii of the input disks against their codes $z \in \mathbb{R}$. The autoencoder appears to represent the disks with their area.}
\label{fig:disk_interpolation}
\end{figure*}

\subsection{Decoding a disk}
\label{subsec:diskDecoding}
\begin{figure*}
\def \disk_no_bias_width{0.19}
\begin{tabularx}{\textwidth}{XXXXX}
\multicolumn{5}{c}{Input}\\
\includegraphics[width=\disk_no_bias_width \textwidth]{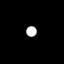}&
\includegraphics[width=\disk_no_bias_width \textwidth]{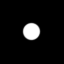}&
\includegraphics[width=\disk_no_bias_width \textwidth]{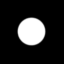}&
\includegraphics[width=\disk_no_bias_width \textwidth]{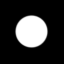}&
\includegraphics[width=\disk_no_bias_width \textwidth]{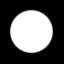}\\
\includegraphics[width=\disk_no_bias_width \textwidth]{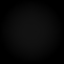}&
\includegraphics[width=\disk_no_bias_width \textwidth]{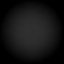}&
\includegraphics[width=\disk_no_bias_width \textwidth]{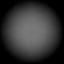}&
\includegraphics[width=\disk_no_bias_width \textwidth]{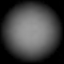}&
\includegraphics[width=\disk_no_bias_width \textwidth]{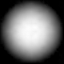}\\
\multicolumn{5}{c}{Output}\\
\\
\includegraphics[width=\disk_no_bias_width \textwidth]{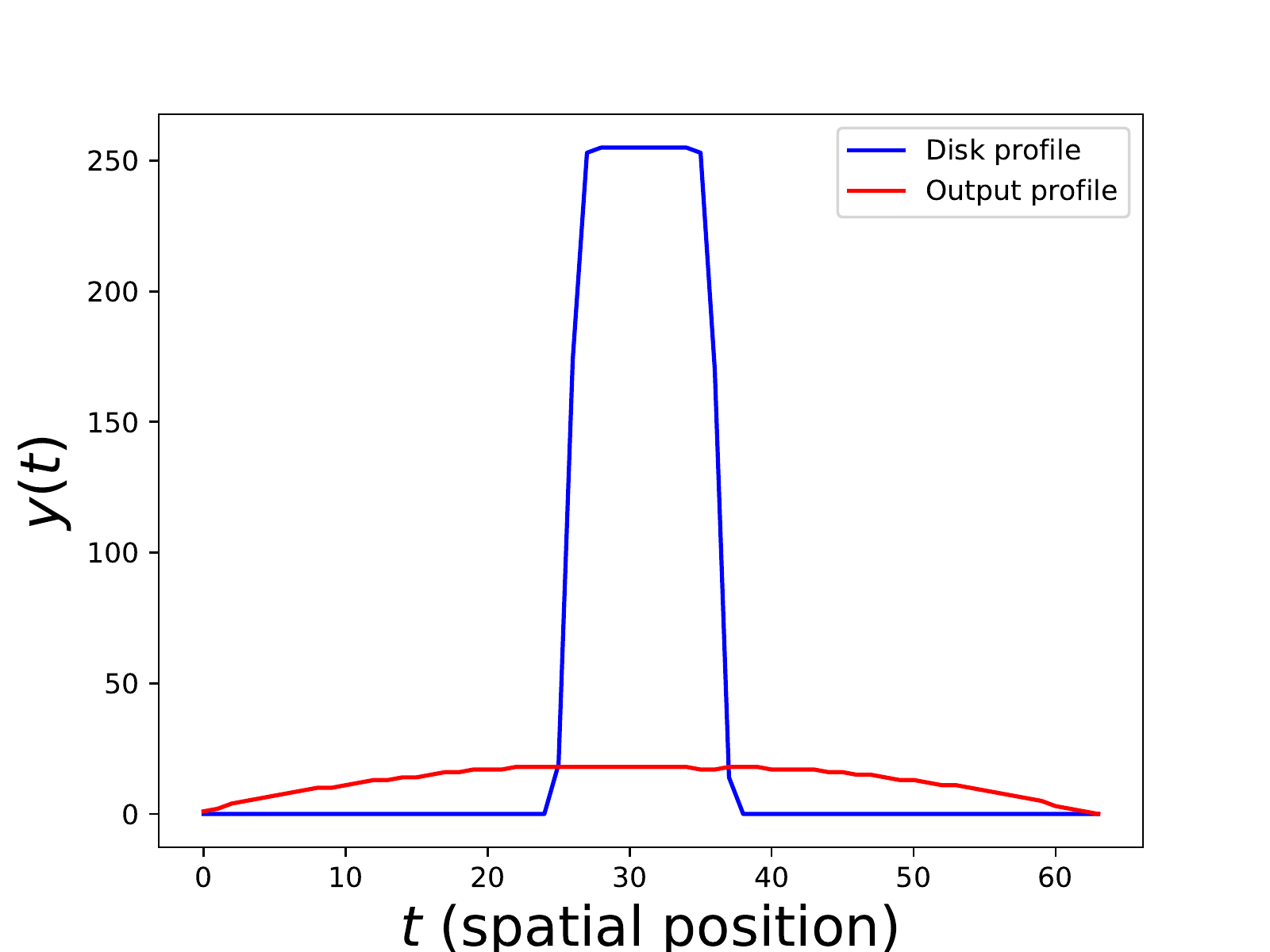}&
\includegraphics[width=\disk_no_bias_width \textwidth]{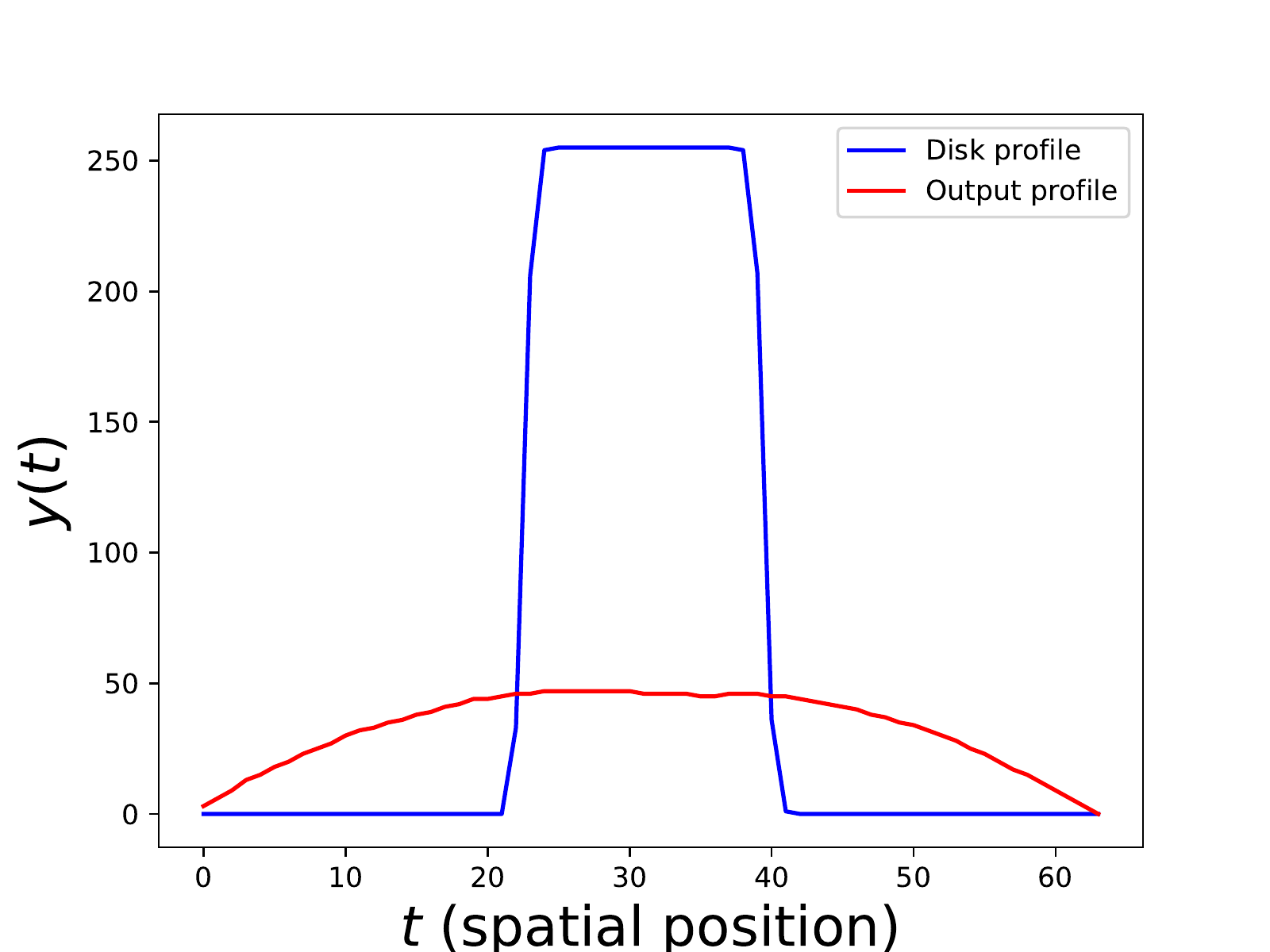}&
\includegraphics[width=\disk_no_bias_width \textwidth]{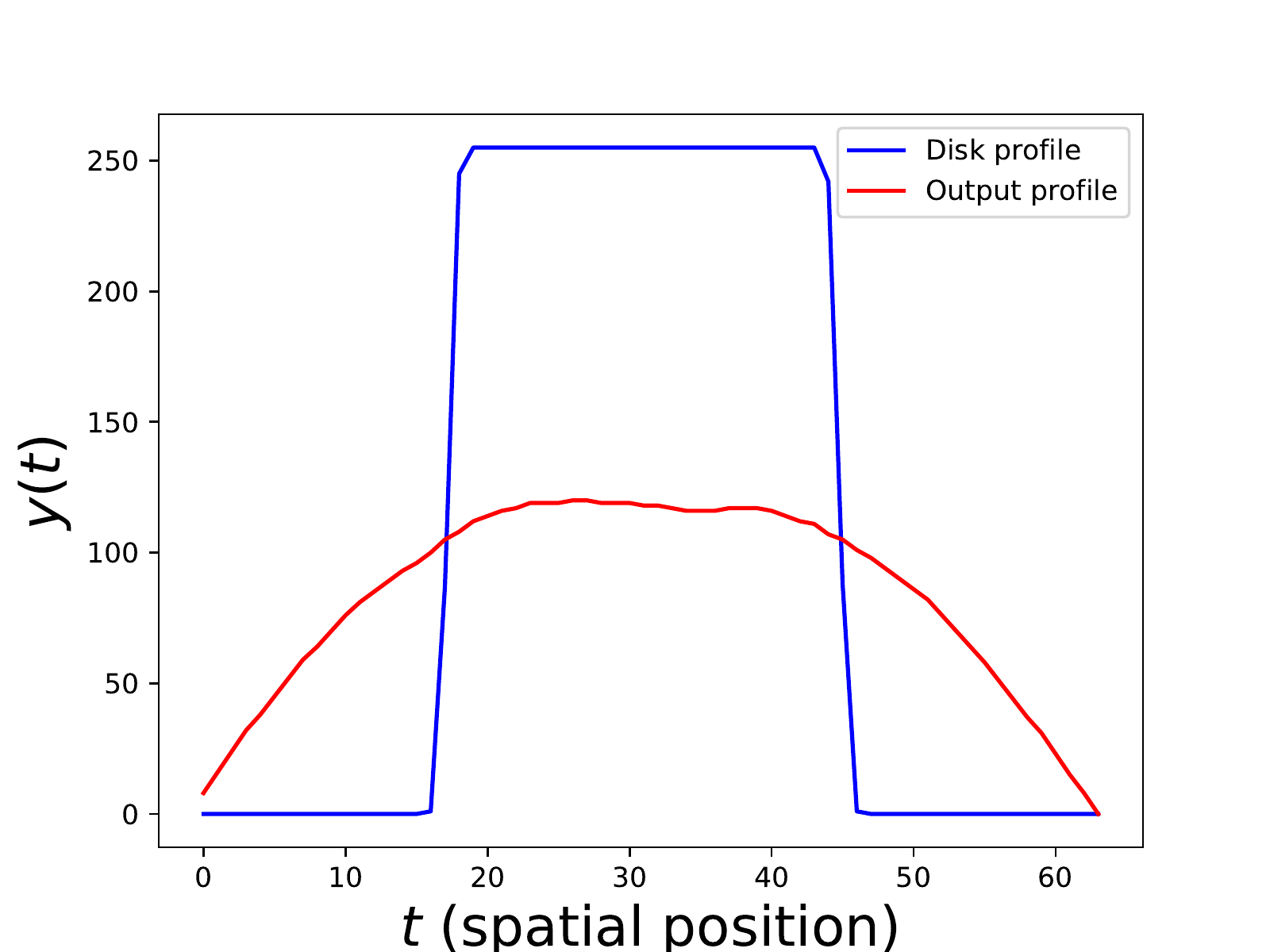}&
\includegraphics[width=\disk_no_bias_width \textwidth]{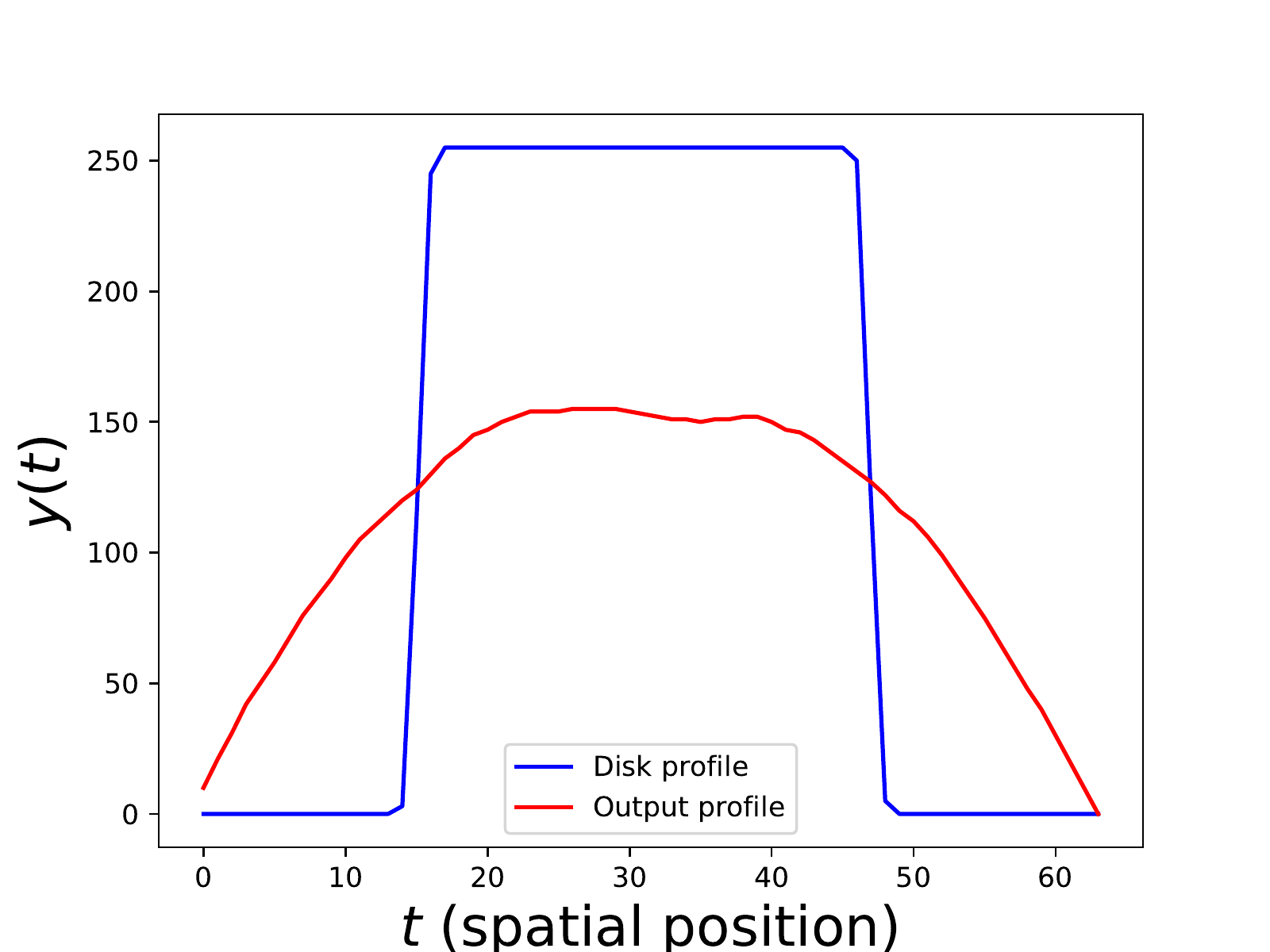}&
\includegraphics[width=\disk_no_bias_width \textwidth]{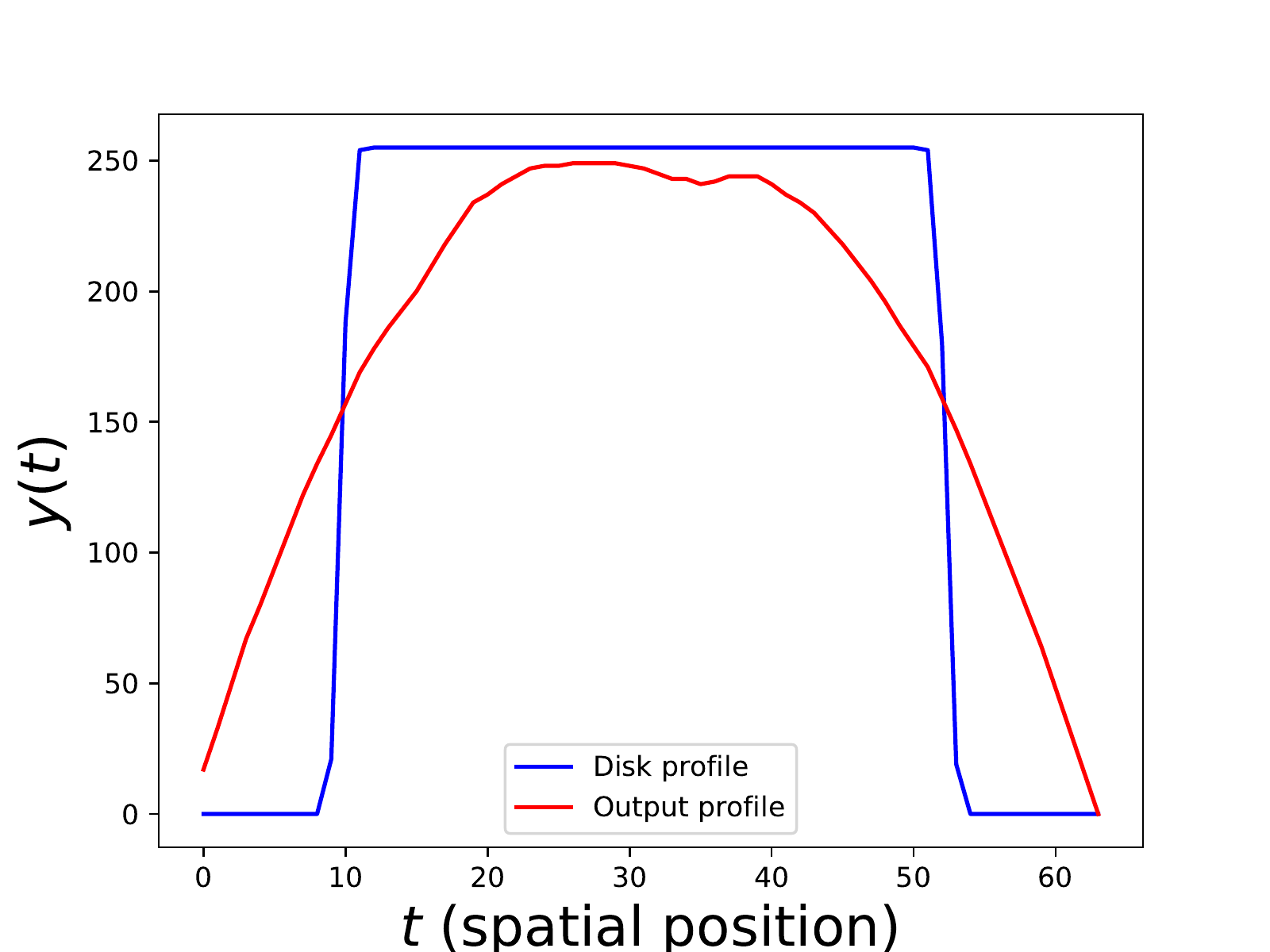}\\
\multicolumn{5}{c}{Profile}\\
\end{tabularx}
\caption{\textbf{Autoencoding of disks when the autoencoder is trained with no bias.} The autoencoder learns a function $f$ which is multiplied by a constant scalar, $h(r)$, for each radius. This behaviour is formalised in Equation~\eqref{eq:noBiasProfileHypothesis}.}
\label{fig:disk_interpolation_no_bias}
\end{figure*}

A more difficult question is how does the autoencoder convert a scalar, $z$, to an output disk of a certain size (the decoding process). One approach to understanding the inner workings of autoencoders, and indeed any neural network, is to remove certain elements of the network and to see how it responds, otherwise known as an \emph{ablation} study. We found that removing the \emph{biases} of the autoencoder leads to very interesting observations. While, as we have shown, the encoder is perfectly able to function without these biases, this is not the case for the decoder. Figure~\ref{fig:disk_interpolation_no_bias} shows the results of this ablation. The decoder learns to spread the energy of $z$ in the output according to a certain function $g$. Thus, the goal of the biases is to shift the intermediary (hidden layer) images such that a cut-off can be carried out to create a satisfactory decoding.

In order to analyse the inner mechanism of the decoder in more depth, we have investigated the behaviour of the decoder in this ablated case (without biases), where it is possible to describe the decoding process with great precision. In particular, we will derive an explicit form for the energy minimized by the network, for which a closed form solution can be found (see Appendix~\ref{app:diskDecoding}), but more importantly for which we will show experimentally that the network indeed finds this solution. We first make a general observation about this configuration (without biases).

\begin{prop}\label{prop:lindecode}[Positive Multiplicative Action of the Decoder Without Bias]

Consider a decoder, without biases $D(z) = D^L \circ \dots \circ D^{1} (z)$, with $D^{\ell+1} = \phi_\alpha\left(U(D^{\ell}) \ast w'_{\ell_i}\right) $, where $U$ stands for upsampling with zero-padding. In this case, the decoder acts multiplicatively on $z$, meaning that 
$$
\forall z,\, \forall \lambda\in\R^+, \: D(\lambda z)=\lambda D(z).
$$
\end{prop}
\proof
{
\noindent Proof :
For a fixed $z$ and for any $\lambda>0$. We have
\begin{alignat}{1}
D^{1} (\lambda z)&= \phi_\alpha\left(U(\lambda z) \ast w'_\ell\right) \notag \\
~&=  \max\left(\lambda (U(z) \ast w'_\ell),0\right) + \alpha \min\left(\lambda(U(z) \ast w'_\ell),0\right)\notag\\
~& =  \lambda \max\left(U(z) \ast w'_\ell,0\right) + \lambda\alpha \min\left(U(z) \ast w'_\ell,0\right) =\lambda \phi_\alpha\left(U(z) \ast w'_\ell\right) = \lambda D^{1}(z).
\end{alignat}
This reasoning can be applied successively to each layer up to the output $y$.
When the code $z$ is one dimensional, the decoder can be summarized as two linear functions, one for positive codes and a second one for the negative codes. However, in all our experiments, the autoencoder without bias has chosen to use only one possible sign for the code, resulting in a linear decoder. \qed
}

\def \disk_width{0.15}
\begin{figure*}
\begin{tabularx}{\textwidth}{XXXXXX}
\multicolumn{6}{c}{Input}\\
\includegraphics[width=\disk_width \textwidth]{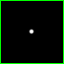}&
\includegraphics[width=\disk_width \textwidth]{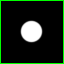}&
\includegraphics[width=\disk_width \textwidth]{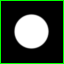}&
\includegraphics[width=\disk_width \textwidth]{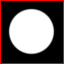}&
\includegraphics[width=\disk_width \textwidth]{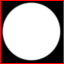}&
\includegraphics[width=\disk_width \textwidth]{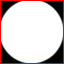}\\
\includegraphics[width=\disk_width \textwidth]{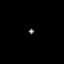}&
\includegraphics[width=\disk_width \textwidth]{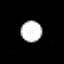}&
\includegraphics[width=\disk_width \textwidth]{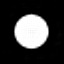}&
\includegraphics[width=\disk_width \textwidth]{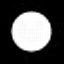}&
\includegraphics[width=\disk_width \textwidth]{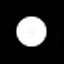}&
\includegraphics[width=\disk_width \textwidth]{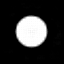}\\
\multicolumn{6}{c}{Output}
\end{tabularx}
\caption{\textbf{Autoencoding of disks with a database with limited radii.} The autoencoder is not able to extrapolate further than the largest observed radius. The images with a green border represent disks whose radii have been observed during training, while those in red have not been observed.}
\label{fig:disk_extrapolation}
\end{figure*}

Furthermore, the profiles in Figure~\ref{fig:disk_interpolation_no_bias} suggest that a single function is learned, and that this function is multiplied by a factor depending on the radius.  In light of Proposition \ref{prop:lindecode}, this means that the decoder has chosen a fixed sign for the code and that the decoder is linear. This can be expressed as
\begin{equation}
D(E( \indic{B_r})) (t) = h(r) f(t),
\label{eq:noBiasProfileHypothesis}
\end{equation}
where $t$ is a spatial variable and $r \in(0,\frac{m}{2}]$ is the radius of the disk. This is checked experimentally in Figure~\ref{fig:noBiasHypothesisVerification} in Appendix~\ref{app:diskDecoding}. In this case, we can write the optimisation problem of the decoder as
\begin{equation}
\hat{f},\hat{h} = \argmin_{f,h} \int_0^R \int_{\Omega} \left( h(r) f(t) - \indic{B_r}(t) \right)^2 dt \; dr,
\label{eq:diskDecodingEnergy}
\end{equation}
where $R$ is the maximum radius observed in the training set, $\Omega=[0,m-1] \times [0,m-1]$ is the image domain, and $B_r$ is the disk of radius $r$. Note that we have expressed the minimisation problem for continuous functions $f$. In this case, we have the following proposition.
\begin{prop}[Decoding Energy for an autoencoder without Biases]
The decoding training problem of the autoencoder without biases has an optimal solution $\hat{f}$ that is radially symmetric and maximises the following energy:
\begin{equation}
J(f) := \int_0^R \left( \int_0^r f(\rho)  \indic{[0,r]}(\rho)\;\rho \;d \rho \right) ^2 \; dr,
\label{eq:diskDecodingEnergyFinal}
\end{equation}
under the (arbitrary) normalization $\lVert f \rVert^2_2 = 1$.
\end{prop}
\proof{
\noindent Proof :
When $f$ is fixed, the optimal $h$ for Equation~\eqref{eq:diskDecodingEnergy} is given by
\begin{equation}
\hat{h}(r) = \frac{\left< f , \indic{B_r} \right>}{ \lVert f \rVert^2_2},
\end{equation}
where $\left< f, \indic{B_r} \right>=\int_\Omega f(t) \indic{B_r} (t) \; dt$. After replacing this in Equation~\eqref{eq:diskDecodingEnergy}, we find that
\begin{alignat}{1}
\hat{f} = \argmin_{f} &\int_0^R - \frac{ \left< f, \indic{B_r}\right>^2}{ \lVert f \rVert^2} dr =
\argmin_{f} \int_0^R - \left< f, \indic{B_r} \right>_2^2 dr,
\end{alignat}
where we have chosen the arbitrary normalisation $\lVert f \rVert^2_2 = 1$.
The form of the last equation shows that the optimal solution is obviously radially symmetric\footnote{If not, then consider its mean on every circle, which decreases the $L^2$ norm of $f$ while maintaining the scalar product with any disk. We then can increase back the energy by deviding by this smaller $L^2$ norm according to $\|f\|_2=1$.}. Therefore, after a change of variables, the energy maximised by the decoder can be written as
\begin{equation}
\int_0^R \left( \int_0^r f(\rho)  \indic{[0,r]}(\rho)\;\rho \;d \rho \right) ^2 \; dr =: J(f),
\end{equation}
such that $\lVert f \rVert^2_2 = 1$. \qed
}

\begin{figure*}
\def \disk_extrapolation_width{0.24}
\begin{tabularx}{\textwidth}{XXXX}
\multicolumn{4}{c}{Input} \\
\includegraphics[width=\disk_extrapolation_width \textwidth]{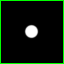}&
\includegraphics[width=\disk_extrapolation_width \textwidth]{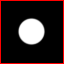}&
\includegraphics[width=\disk_extrapolation_width \textwidth]{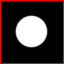}&
\includegraphics[width=\disk_extrapolation_width \textwidth]{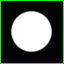}\\
\includegraphics[width=\disk_extrapolation_width \textwidth]{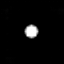}&
\includegraphics[width=\disk_extrapolation_width \textwidth]{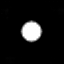}&
\includegraphics[width=\disk_extrapolation_width \textwidth]{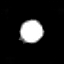}&
\includegraphics[width=\disk_extrapolation_width \textwidth]{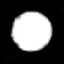}
\end{tabularx}
\caption{\textbf{Input and output of our network when autoencoding examples of disks when the database contains a ``hole''.} Disks of radii between 11 and 18 pixels (out of 32) were not observed in the database. In green, the disks whose radii have been observed in the database, in red those which have not.}
\label{fig:diskInterpolationFailCodeSize}
\end{figure*}

In Appendix~\ref{app:diskDecoding}, we compare the numerical solution of this problem with the actual profile learned by the network, yielding a very close match. This result is enlightening, since it shows that the training process has achieved the optimal solution, in spite of the fact that the loss is non convex.


\subsection{Generalisation and regularisation}
\label{subec:regularisation}

As we have recalled in Section\ref{sec:prior}, many works have recently investigated the generative capacity of autoencoders or GANs. Nevertheless, it is not clear that these architectures truly invent or generalize some visual content. A simpler question is : to what extent is the network able to generalise in the case of the simple geometric notion of size ? In this section, we address this issue in our restricted but interpretable case.

For this, we study the behaviour of our autoencoder when examples are removed from the training dataset. In Figure~\ref{fig:disk_extrapolation}, we show the autoencoder result when the disks with radii above a certain threshold $R$ are removed. The radii of the left three images (with a green border) are present in the training database, whereas the radii of the right three (red border) have not been observed. It is clear that the network lacks the capacity to extrapolate further than the radius $R$. Indeed, the autoencoder seems to project these disks onto smaller, observed, disks, rather than learning the abstraction of a disk.

Again by removing the biases from the network, we may explain why the autoencoder fails to extrapolate when a maximum radius $R$ is imposed. In Appendix~\ref{app:diskDecodingExtrapolationFailure}, we show experimental evidence that in this situation, the autoencoder learns a function $f$ whose support is restricted by the value of $R$, leading to the autoencoder's failure. However, a fair criticism of the previous experiment is simply that the network (and deep learning in general) is not designed to work on data which lie outside of the domain observed in the training data set. Nevertheless, it is reasonable to expect the network to be robust to holes \emph{inside} the domain. Therefore, we have also analysed the behaviour of the autoencoder when we removed training datapoints whose disks' radii lie within a certain range, between 11 and 18 pixels (out of a total of 32). We then attempt to reconstruct these points in the test data. Figure~\ref{fig:diskInterpolationFailCodeSize} shows the results of this experiment failure. Once again, in the unknown regions the network is unable to recreate the input disks. Several explanations in the deep learning literature of this phenomenon, such as a high curvature of the underlying data manifold \cite{Goodfellow2016Deep} (see page 521, or end of Section 14.6), noisy data or high intrinsic dimensionality of the data \cite{Bengio2005Non}. In our setting, \emph{none of these explanations is sufficient}. Thus we conclude that, even in the simple setting of disks, the classic autoencoder cannot generalise correctly when a database contains holes. 

Consequently, this effect is clearly due to the gap between two different formulations of the loss of an autoencoder :
\begin{align}
\mathcal{L}_1 &=\Expect_{x \sim p_x} \lVert x - D(E(x)) \rVert^2 \\
\mathcal{L}_2 &=\Expect_{x \in \mathrm{dataset}} \lVert x - D(E(x)) \rVert^2.
\end{align}
The latter supposes that the dataset faithfully reflects the distribution $p_x$ of images and is the empirical loss actually used in most of the literature. In our setting we are able to faithfully sample the true distribution $p_x$ and study what happens when a certain part of the distribution is not well observed.

This behaviour is potentially problematic for applications which deal with more complex natural images, lying on a high-dimensional manifold, as these are very likely to contain such holes. We have therefore carried out the same experiments using the recent DCGAN approach of~\cite{radford2015unsupervised}. The visual results of their algorithm are displayed in Appendix~\ref{app:autoencodingDCGAN}. We trained their network using a code size of $d=1$ in order to ensure fair comparisons. The network fails to correctly autoencode the disks belonging to the unobserved region. Indeed, \emph{GAN-type networks may not be very good at generalising data}, since their goal is to find a way to map the observed data to some predefined distribution, therefore there is no way to modify the latent space itself.
This shows that the generalisation problem is likely to be ubiquitous, and indeed observed in more sophisticated networks, designed to learn natural images manifolds, even in the simple case of disks. We therefore believe that this issue deserves careful attention. Actually this experiment suggets that the capacity to generate new and simple geometrical shapes could be taken as a minimal requirement for a given architecture. 

In order to address the problem, we now investigate several regularisation techniques whose goal is to aid the generalisation capacity of neural networks.
\subsubsection{Regularisation}

\begin{figure*}
\def \disk_extrapolation_width{0.24}
\begin{tabularx}{\textwidth}{@{}XXXX@{}}
\multicolumn{4}{c}{Disk radius $r$ as a function of the latent code $z$}\\
\includegraphics[width=\disk_extrapolation_width \textwidth]{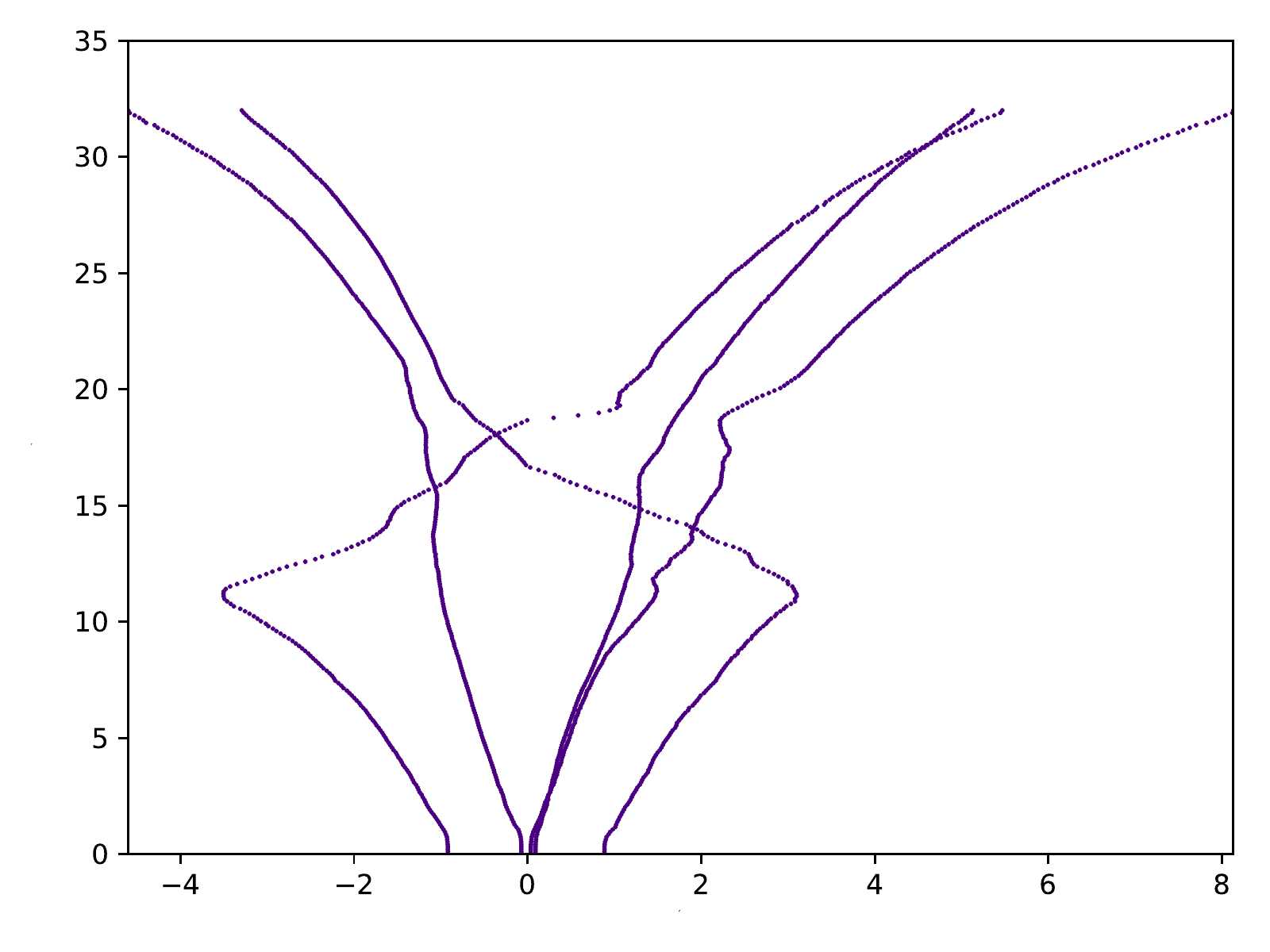}&
\includegraphics[width=\disk_extrapolation_width \textwidth]{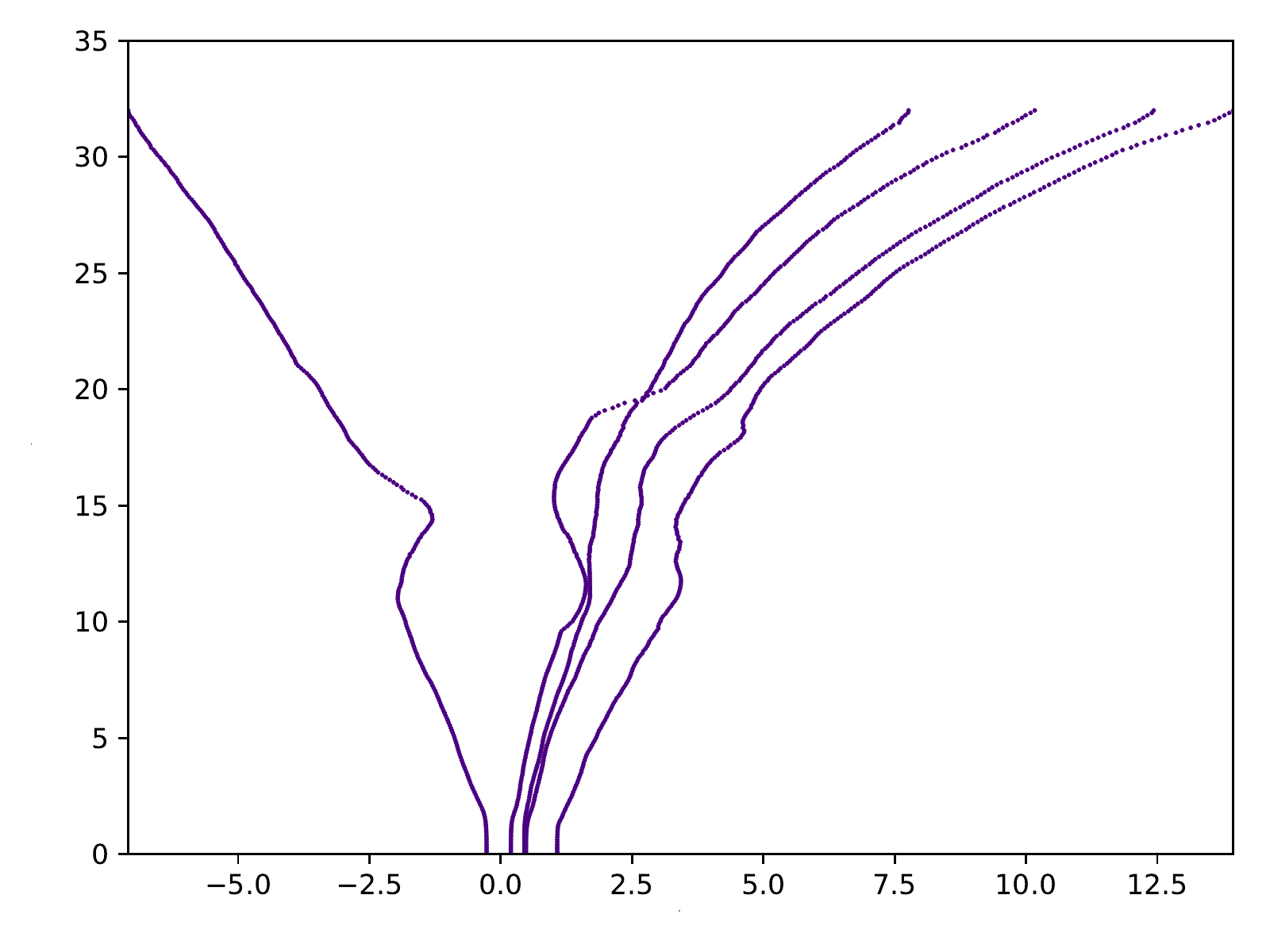}&
\includegraphics[width=\disk_extrapolation_width \textwidth]{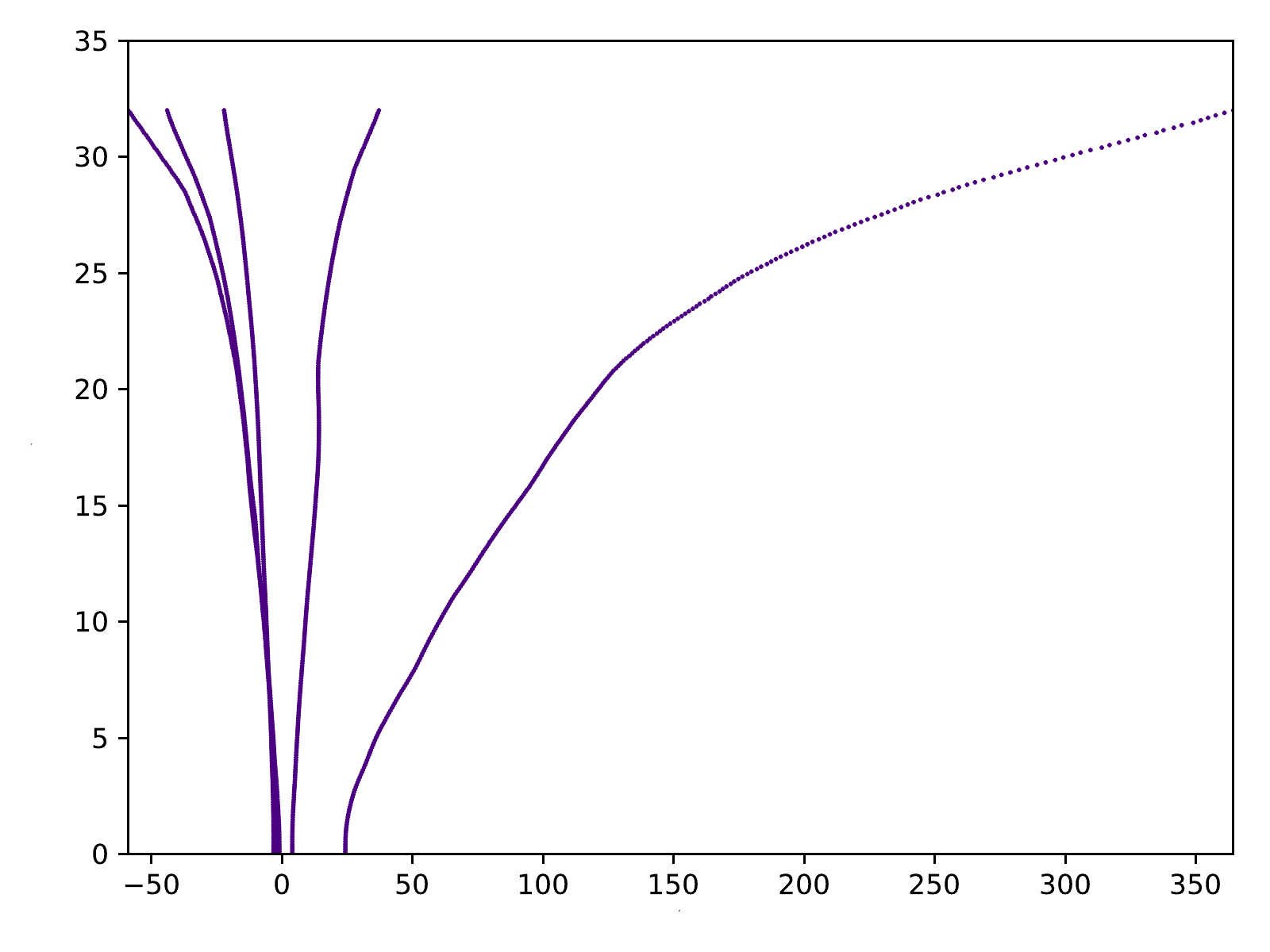}&
\includegraphics[width=\disk_extrapolation_width \textwidth]{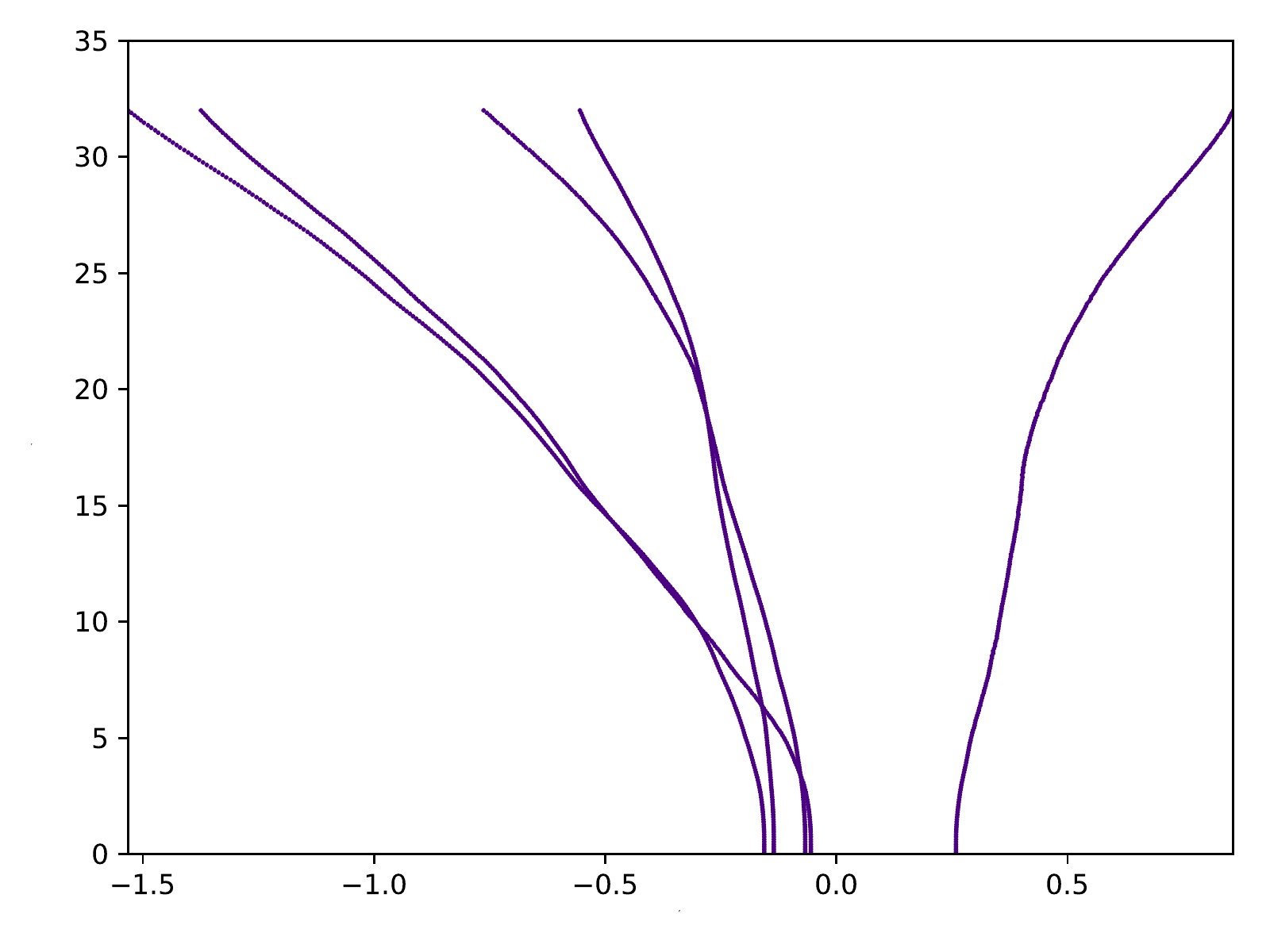}\\
\centering No regularisation & \centering $\psi_1$ &\centering $\psi_2$ & \centering $\psi_3$ \tabularnewline
\end{tabularx}

\vspace{4mm}
\def \disk_interpolation_width{0.2}
\begin{tabularx}{\textwidth}{@{}XXXXX@{}}
\hline
\multicolumn{5}{c}{Autoencoder output}\\
\includegraphics[width=\disk_interpolation_width \textwidth]{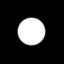}&
\includegraphics[width=\disk_interpolation_width \textwidth]{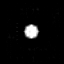}&
\includegraphics[width=\disk_interpolation_width \textwidth]{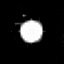}&
\includegraphics[width=\disk_interpolation_width \textwidth]{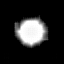}&
\includegraphics[width=\disk_interpolation_width \textwidth]{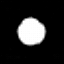}\\
\centering Input & \centering No regularisation & \centering $\psi_1$ & \centering $\psi_2$ & \centering $\psi_3$ \tabularnewline
\end{tabularx}
\caption{\textbf{Result of different types of regularisation on autoencoding in an ``unknown region'' of the training database.} We have encoded/decoded a disk which was not observed in the training dataset. We show the results of four experiments: no regularisation, $\ell_2$ regularisation in the latent space ($\psi_1$), $\ell_2$ weight penalisation of the encoder and decoder ($\psi_2$) and $\ell_2$ weight penalisation of the encoder only ($\psi_3$). In order to highlight the instability of the autoencoder without regularisation, we have carried out the same experiment five times, and shown the resulting latent spaces for each experiment. The latent spaces produced by a regularised autoencoder, and in particular types 2-3, are consistently smoother than the unregularised version, which can produce incoherent latent spaces, and thus incorrect outputs.}
\label{fig:diskInterpolationRegularisation}
\end{figure*}

\begin{figure*}
\def \disk_interpolation_width{0.19}
\begin{tabularx}{\textwidth}{@{}XXXXX@{}}
\multicolumn{5}{c}{Input}\\
\includegraphics[width=\disk_interpolation_width \textwidth]{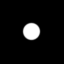}&
\includegraphics[width=\disk_interpolation_width \textwidth]{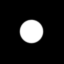}&
\includegraphics[width=\disk_interpolation_width \textwidth]{\imageDir disk_regularisation/regularisation_0/disk_number_000130_radius_000014_patch_in.png}&
\includegraphics[width=\disk_interpolation_width \textwidth]{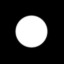}
\\
\\
\multicolumn{5}{c}{Regularisation type 3, $\lambda=0.1$}
\\
\includegraphics[width=\disk_interpolation_width \textwidth]{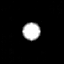}&
\includegraphics[width=\disk_interpolation_width \textwidth]{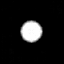}&
\includegraphics[width=\disk_interpolation_width \textwidth]{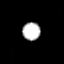}&
\includegraphics[width=\disk_interpolation_width \textwidth]{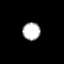}&
\includegraphics[width=\disk_interpolation_width \textwidth]{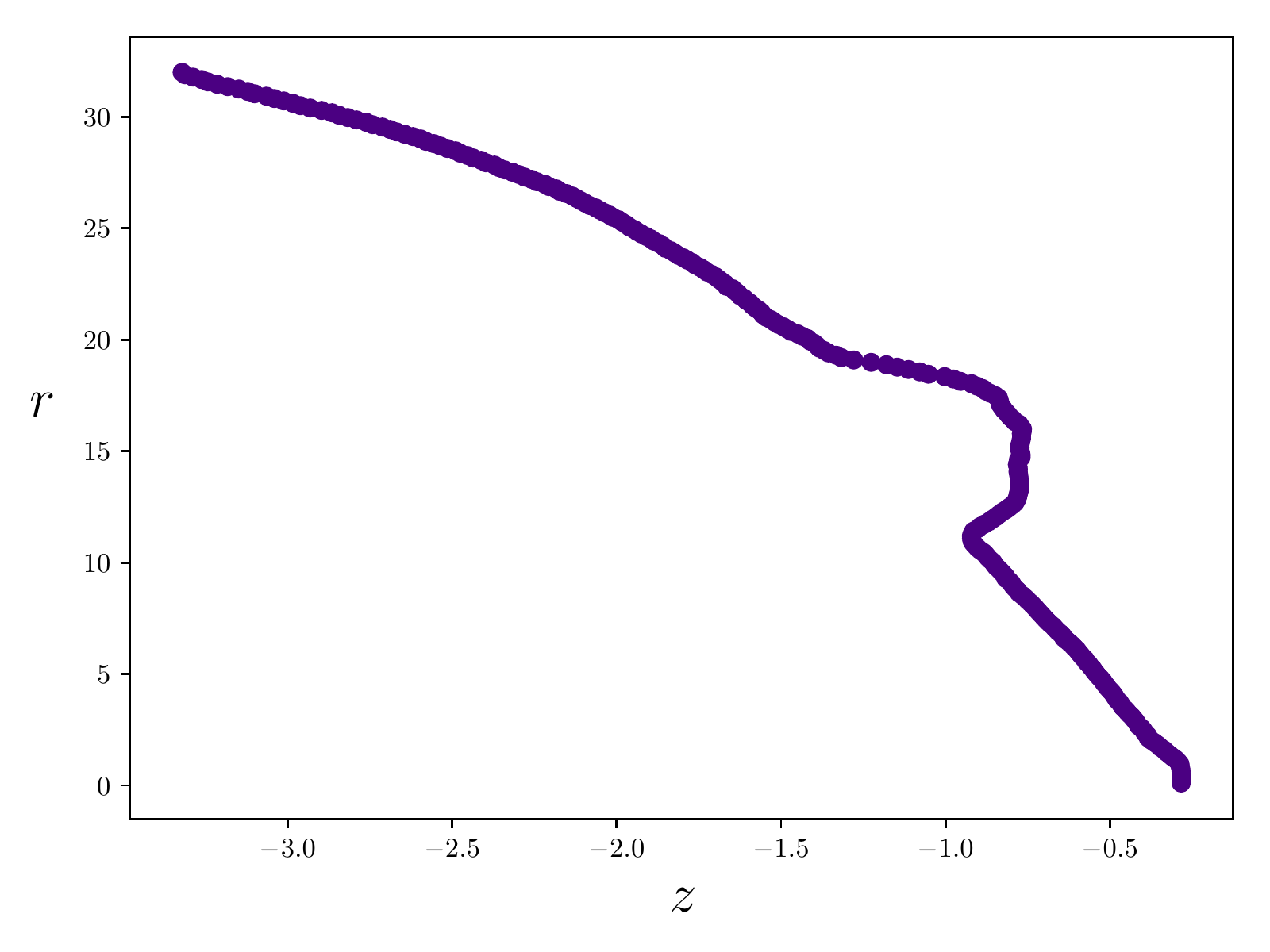}
\\
\multicolumn{5}{c}{Regularisation type 3, $\lambda=1.0$}
\\
\includegraphics[width=\disk_interpolation_width \textwidth]{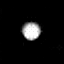}&
\includegraphics[width=\disk_interpolation_width \textwidth]{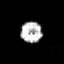}&
\includegraphics[width=\disk_interpolation_width \textwidth]{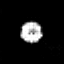}&
\includegraphics[width=\disk_interpolation_width \textwidth]{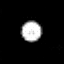}&
\includegraphics[width=\disk_interpolation_width \textwidth]{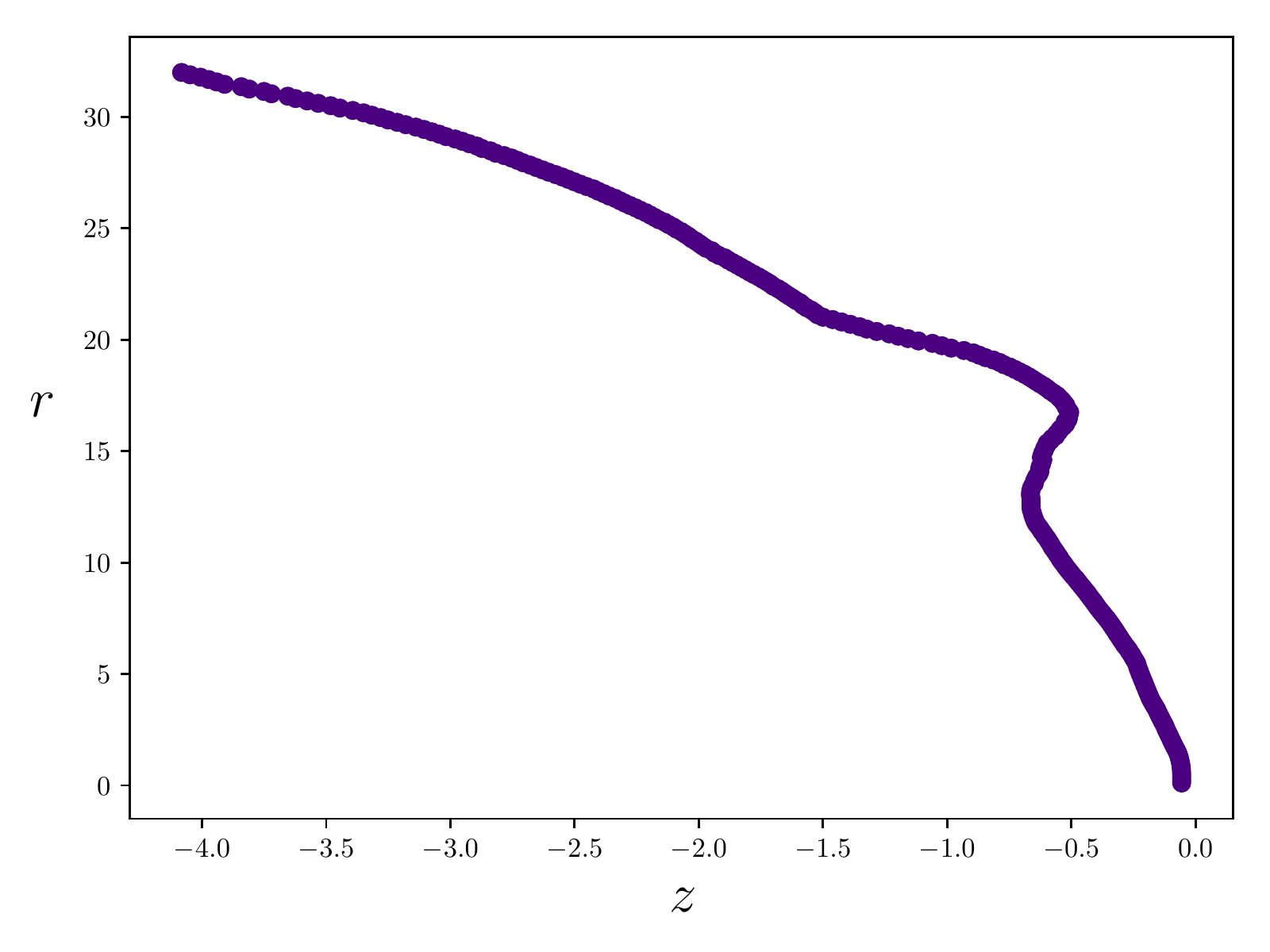}
\\
\multicolumn{5}{c}{Regularisation type 3, $\lambda=10.0$}
\\
\includegraphics[width=\disk_interpolation_width \textwidth]{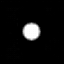}&
\includegraphics[width=\disk_interpolation_width \textwidth]{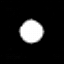}&
\includegraphics[width=\disk_interpolation_width \textwidth]{\imageDir disk_regularisation/regularisation_3/lambda_10.0/disk_number_000130_radius_000014_patch_out_disk_missing.png}&
\includegraphics[width=\disk_interpolation_width \textwidth]{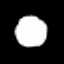}&
\includegraphics[width=\disk_interpolation_width \textwidth]{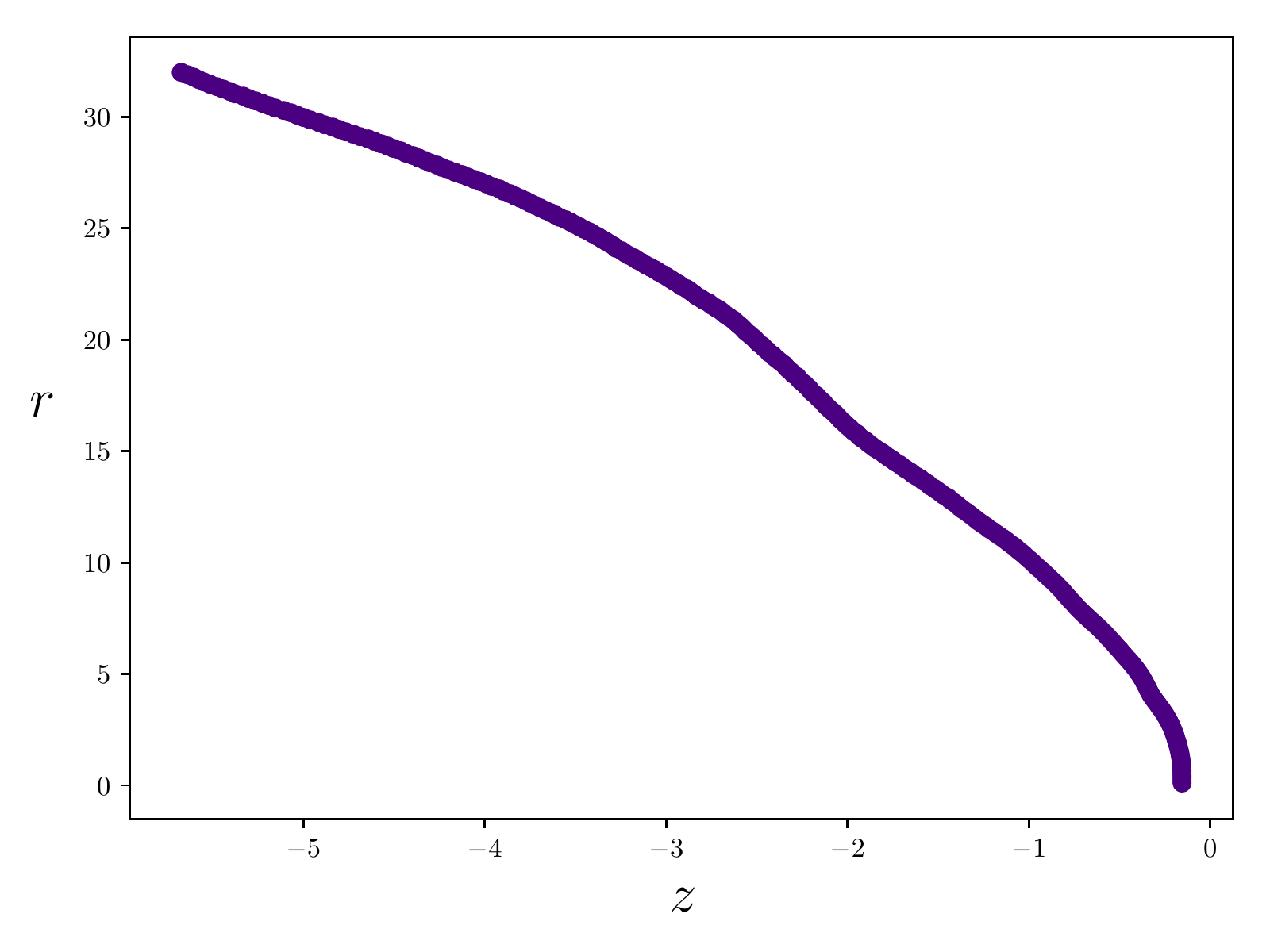}
\\
\multicolumn{5}{c}{Regularisation type 3, $\lambda=50.0$}
\\
\includegraphics[width=\disk_interpolation_width \textwidth]{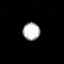}&
\includegraphics[width=\disk_interpolation_width \textwidth]{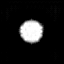}&
\includegraphics[width=\disk_interpolation_width \textwidth]{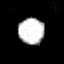}&
\includegraphics[width=\disk_interpolation_width \textwidth]{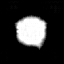}&
\includegraphics[width=\disk_interpolation_width \textwidth]{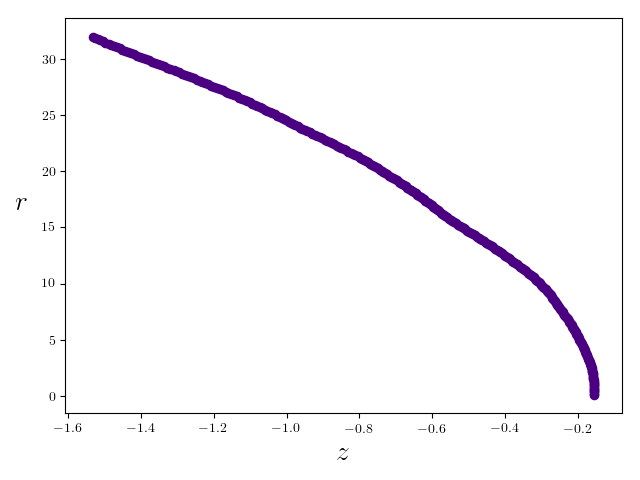}
\\
\multicolumn{5}{c}{Regularisation type 3, $\lambda=100.0$}
\\
\includegraphics[width=\disk_interpolation_width \textwidth]{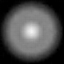}&
\includegraphics[width=\disk_interpolation_width \textwidth]{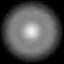}&
\includegraphics[width=\disk_interpolation_width \textwidth]{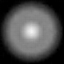}&
\includegraphics[width=\disk_interpolation_width \textwidth]{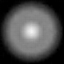}&
\includegraphics[width=\disk_interpolation_width \textwidth]{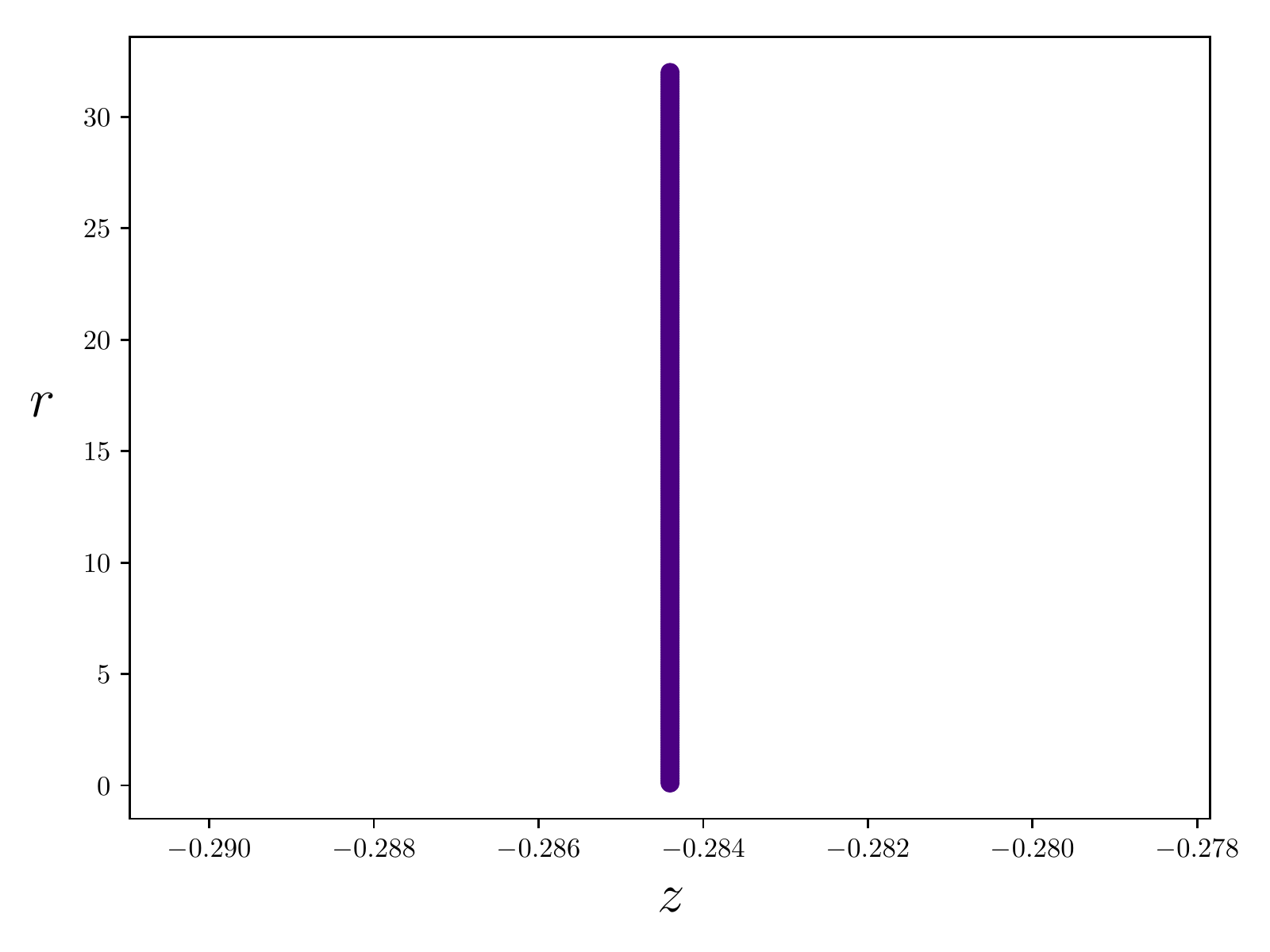}
\\
\end{tabularx}
\caption{\textbf{Effect of encoder regularisation on the generalisation capacity of the network}. Regularisation of the network with a varying value of $\lambda$, using the regularisation $\psi_3$ (encoder regularisation) described in Section~\ref{subec:regularisation}}
\label{fig:diskInterpolationRegularisationtTypeThree}
\end{figure*}

We would like to impose some structure on the latent space in order to interpolate correctly in the case of missing datapoints. This is often achieved via some sort of regularisation. This regularisation can come in many forms, such as imposing a certain distribution in the latent space, as in variational autoencoders (\cite{Kingma2014Auto}), or by encouraging $z$ to be sparse, as in sparse auto-encoders (\cite{Ranzato2007Sparse,Makhzani2013K}). In the present case, the former is not particularly useful, since a probabilistic approach will not encourage the latent space to correctly interpolate. The latter regularisation does not apply, since we already have $d=1$. Another commonly used approach is to impose an $\ell_2$ penalisation of the weights of the filters in the network. The idea behind this bears some similarity to sparse regularisation; we wish for the latent space to be as simple as possible, and therefore hope to avoid over-fitting.

We have implemented several regularisation techniques on our network. Firstly, we attempt a simple regularisation of the latent space by requiring a locality-preservation property as suggested in \cite{Hadsell2006Dimensionality,alain2014regularized,liao2017graph}, namely that the $\ell_2$ distance  between two images ($x$,$x'$) be maintained in the latent space. This is done by randomly selecting a neighbour of each element in the training batch. Secondly, we regularise the weights of the encoder and/or the decoder (also known as weight decay).
Our training attempts to minimise the sum of the data term, $\lVert x - D(E(x)) \rVert^2_2$, and a regularisation term $\lambda \psi(x,\theta)$, which can take one of the following forms:
\vspace{-3mm}
\begin{itemize}
    \setlength\itemsep{0.5em}
	\item Type 1 : $ \psi_1(x,x') = ( \lVert x - x'\rVert^2_2 - \lVert E(x) - E(x')\rVert^2_2  )^2$
	\item Type 2 : $ \psi_2(\Theta_E,\Theta_D) = \sum_{\ell=1}^{L}  \lVert w_{\cdot,\ell} \rVert^2_2 + \lVert w'_{\cdot,\ell} \rVert^2_2 $
	\item Type 3 : $ \psi_3(\Theta_E) = \sum_{\ell=1}^{L}  \lVert w_{\cdot,\ell} \rVert^2_2 $.
\end{itemize}
\vspace{-3mm}
We note here that, given the very strong bottleneck of our architecture, the dropout regularisation technique does not make much sense here. 

Figure~\ref{fig:diskInterpolationRegularisation} shows the results of these experiments. First of all, we observe that $\psi_1$ does not work satisfactorily. One interpretation of this is that the manifold in the training data is ``discontinuous'', and therefore there are no close neighbours for the disks on the edge of the unobserved region. Therefore, this regularisation is to be avoided in cases where there are significant holes in the sampling of the data manifold. The second type of regularisation, minimising the $\ell_2$ norm of the encoder and decoder weights, produces a latent space which appears smooth, however the final result is not of great quality. Finally, we observe that regularising the weights of the encoder ($\psi_3$) works particularly well, and that the resulting manifold is smooth and correctly represents the area of the disks. Consequently, this asymmetrical regularisation approach is to be encouraged in other applications of autoencoders. We show further results of this regularisation approach in Figure~\ref{fig:diskInterpolationRegularisationtTypeThree}, when the regularisation parameter is varied. We see that increasing this parameter smooths the latent space, until $\lambda$ becomes too great and the training fails.


At this point, we take the opportunity to note that the clear, marked effects seen with the different regularisation approaches are consistently observed in different training runs. This is due in large part to the controlled, simple setting of autoencoding with disks. Indeed, many other more sophisticated networks, especially GANs, are known to be very difficult to train~\cite{salimans2016improved}, leading to unstable results or poor reproducibility. We hope that our approach can be of use to more high-level applications, and possibly serve as a sanity check to which these complicated networks should be submitted. Indeed, it is reasonable to assume that such networks should be able to perform well in simple situations before moving onto complicated data

\section{Encoding position in an autoencoder}
\label{sec:encodingPosition}

We now move on to the analysis of our second geometric property : position. For this, we ask the following question : is it possible to encode the position of a simple one-hot vector (a discretised Dirac in other words) to a scalar, and if so, how ? A similar situation was investigated concurrently to our work by Liu et al. \cite{liu2018intriguing}, who studied a network which projected images of randomly positioned squares to a position (a vector in $\mathbb{R}^2$), and then back again to the pixel space, with as small a loss as possible. Their opinion was that this was not possible, at least to a satisfactory degree, by training neural networks, which lead them to propose the CoordConv network layer.

In the following, we hand-craft a simple neural network which can achieve this in the forward direction : from a one-hot vector to the position. To simplify, we will analyse the 1-D case, that is to say the input lives in a one dimensional space. 

Firstly, let us define some notation. We denote $x \in \mathbb{R}^n$ the input to the network, where $n$ is the input dimension. We shall denote with $u^{(\ell)}$ the output of the $\ell$-th layer of the neural network. We shall denote with $\varphi$ the filter of our network. We shall consider the following hand-crafted filter :
\begin{equation}
\varphi = \left[1, 2, 1\right].
\label{eq:handCraftedPositionWeights}
\end{equation}
Let us also suppose that subsampling factor is $s=2$, and that it takes place at every even position (0, 2, 4 etc). We denote with $\mathcal{S}$ the subsampling operator. We do not use any non-linearities or biases in the network. Finally, we denote with $E$ the whole linear neural network.
\subsection{Some concrete examples}
\label{subsec:positionExamples}

As a simple example, let use consider an input vector $x = \left[1,0,0,0\right]$. After the first filtering and subsampling step, we have $u^{(1)} = \left[2,0\right]$, and then $u^{(2)} = 4$. Similarly, if $x = \left[0,0,0,1\right]$, then $u^{(2)} = 1$. Thus, with these two simple operations, it seems we can extract the non-zero position of a one-hot vector.

To take another example of the result of these operations, let us take a look at a similar result in $n=8$. In Table~\ref{tab:eightVector}, we can see the results for every possible 1-hot vector. Indeed, the network seems to extract the position of the one-hot vector.

\begin{table}
    \begin{tabular}{|c|| c c c c|}
        \hline
        $x$ &$[1, 0, 0, 0, 0, 0, 0, 0]$ & $[0, 1, 0, 0, 0, 0, 0, 0]$ & $[0, 0, 1, 0, 0, 0, 0, 0]$ & $[0, 0, 0, 1, 0, 0, 0, 0]$ \\
        $u^{(1)}$ & $[2, 0, 0, 0]$ & $[1, 1, 0, 0]$ & $[0, 2, 0, 0]$ & $[0, 1, 1, 0]$\\
        $u^{(2)}$ & $[4, 0]$ & $[3, 1]$ & $[2, 2]$ & $[1, 3]$\\
        $u^{(3)}$ & $[8]$ & $[7]$ & $[6]$ & $[5]$\\
        \hline
    \end{tabular}

    \begin{tabular}{|c|| c c c c|}
        \hline
        $x$ &$[0, 0, 0, 0, 1, 0, 0, 0]$ & $[0, 0, 0, 0, 0, 1, 0, 0]$ & $[0, 0, 0, 0, 0, 0, 1, 0]$ & $[0, 0, 0, 0, 0, 0, 0, 1]$ \\
        $u^{(1)}$ & $[0, 0, 2, 0]$ & $[0, 0, 1, 1]$ & $[0, 0, 0, 2]$ & $[0, 0, 0, 1]$\\
        $u^{(2)}$ & $[0, 4]$ & $[0, 3]$ & $[0, 2]$ & $[0, 1]$\\
        $u^{(3)}$ & $[4]$ & $[3]$ & $[2]$ & $[1]$\\
        \hline
    \end{tabular}
\caption{Results of all possible one-hot vectors of size eight in the simple linear neural network described in Section~\ref{sec:encodingPosition}}
\label{tab:eightVector}
\end{table}

\subsection{Position encoding in general the case}
\label{subsec:positionTheory}

Now, if we take the general case, $x \in \mathbb{R}^n$ with $n=2^L$ and where $L$ is the total number of layers, then the output of each layer $u^{(\ell)}$ can be written in terms of the convolution with the previous layer :

\begin{equation}
u{^{(\ell)}}(t) = \sum_{i \in \mySet{A}} \varphi(i) u^{(\ell-1)}(s t - i),
\end{equation}
where $\mySet{A}$ is defined as the support of the filter $\varphi$. In our case, $\mySet{A}=\{-1, 0, 1\}$. Using an induction argument, we can show that the network $E$ indeed extracts the position of the one-hot input vector. More precisely, as we have seen in Section~\ref{subsec:positionExamples}, the network extracts the position in an inverted order, that is to say $n-a+1$, if $a$ is the postion of the non-zero element of $x$ and if we number the elements of $x$ from $x_0$ to $x_{2^L-1}$. 

\begin{prop}[The linear neural network $E$ extracts the position of a Dirac input]
\label{prop:positionEncoder}
Consider the neural network $E$ 
described earlier in this section, and a one-hot input vector $x \in \mathbb{R}^n$, with $n=2^L$ and where $(x_i), i \in \left[ 0, \dots, n-1\right]$ denotes the $i^{th}$ element of $x$.
\end{prop}

\def \disk_width{0.2}
\begin{figure*}
\begin{tabularx}{\textwidth}{XXXXXX}
\includegraphics[width=\disk_width \textwidth]{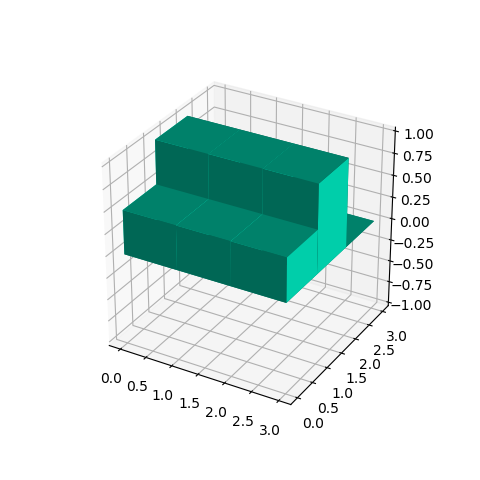}&
\includegraphics[width=\disk_width \textwidth]{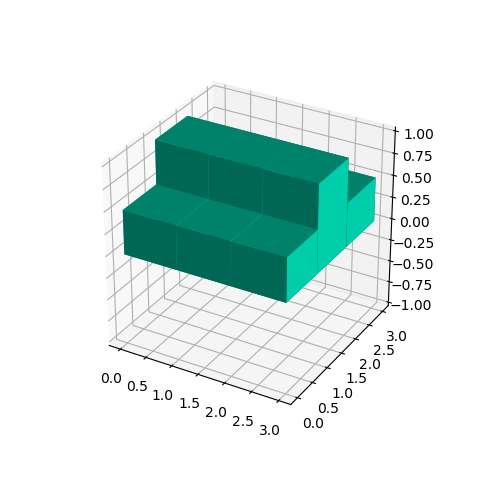}&
\includegraphics[width=\disk_width \textwidth]{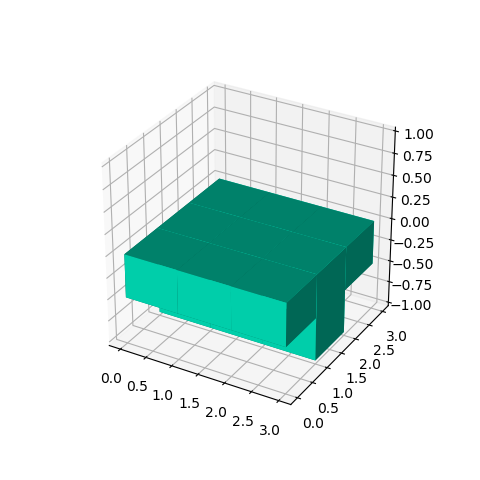}&
\includegraphics[width=\disk_width \textwidth]{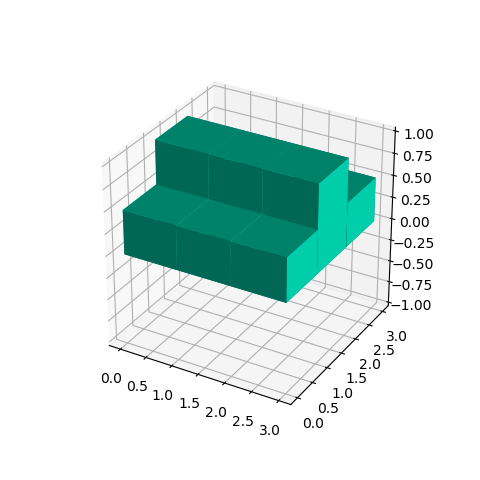}&
\includegraphics[width=\disk_width \textwidth]{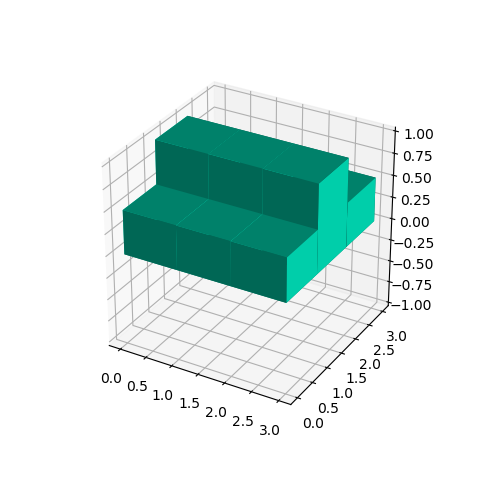}&
\includegraphics[width=\disk_width \textwidth]{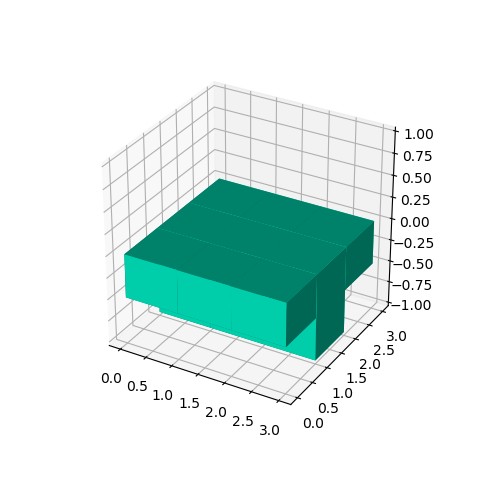}\\
\multicolumn{6}{c}{Weights for each layer of the encoder network $E$}
\end{tabularx}
\caption{\textbf{Weights of position encoder network}. We show the weights found by the encoder network $E$. The 1D weights are shown in a 2D representation for clearer exposition. These weights agree with our theoretical prediction in Section~\ref{subsec:positionTheory}}
\label{fig:position_weights}
\end{figure*}

\proof{
\noindent Proof :

We prove this by induction over the number of layers in the network.

\paragraph{One hidden layer}
This is easy to verify for a network with one hidden layer. Indeed, if the input $x \in \mathbb{R}^2$ contains a 1 at the first ($0^{th}$) position, then the network output is $2*1=2$. If $x$ contains a 1 at the second position, then the network output is $1*1=1$. Thus, the property is true for the case of one hidden layer.

\paragraph{$L$ hidden layers}
Let us suppose that the network contains $L$ hidden layers, and extracts the non-zero position in reverse order, that is to say $u^{(L)} = 2^L-a+1$, where $a$ is the non-zero position in $x$. Since the output of the network is a positive linear combination of the input vector with fixed coefficients, and the property holds for any $a$, we can rewrite the output as

\begin{equation}
\label{eq:inductionFormula}
    E(x) = \sum_{i=0}^{2^L-1} (2^L - i) x_i.
\end{equation}
Now let us suppose that we add a layer above the input layer, so that the network now has $L+1$ hidden layers and the input $x$ now belongs to $\mathbb{R}^{2^{L+1}}$, and the previous $x$ is now $u^{(1)}$. We can determine the output of the network with Equation~\eqref{eq:inductionFormula}. There are three cases to distinguish between.

Suppose first that $a$ is an even position, so that $\exists k \in \mathbb{N} , a=2k$. Thus, using Equation~\eqref{eq:inductionFormula}, we have that
\begin{align}
\label{eq:inductionFirstCase}
    E(x) =& \sum_{i=0}^{2^L-1} (2^L - i) u^{(1)}(i)  \nonumber \\
            = &\; (2^L - k) . 2  \nonumber \\
            = &\; 2^{(L+1)} -2k.
\end{align}
Thus, we find that the network extracts the correct ``inverted-order'' position,  with $a=2k$.

Let us suppose now that $a = 2k+1$. In this case, we have
\begin{align}
\label{eq:inductionSecondCase}
    E(x) = &\; (2^L - k) . 1 + (2^L - (k+1)) . 1 \nonumber \\
            = &\; 2^{(L+1)} - (2k + 1).
\end{align}
Again, the network correctly identifies the position $a=2k+1$.

Finally, there is a special case, where $a = 2^{(L+1)}-1 = 2k+1$, with $k=2^L-1$ (at the end of the vector $x$). In this case, we have
\begin{align}
\label{eq:inductionThirdCase}
    E(x) = &\; (2^L - k) . 1 \nonumber \\
            = &\; 2^L - (2^L -1) \nonumber \\
            = &\; 1.
\end{align}
Thus, in the extreme case of $a = 2^{(L+1)}-1$, $E$ still extracts the inverted-order position. Thus, we have proved that the network $E$ extracts the position $k$ of the non-zero element of a one-hot input vector.
\hfill$\qed$
}

Furthermore, obviously any variant $\varphi' = b \varphi$, with $b \in \mathbb{R}^{*}$ also extracts the position. Finally, we note that our proof relies on the fact that the subsampling factor is $s=2$. Therefore, this appears important to any neural network which wishes to represent position internally.

\subsection{Experimental results}

We now present experimental evidence that training a neural network with the above architecture leads to the hand-crafted weights in practice. We use the neural network described in Section~\ref{sec:aeArchitecture}, where the depths of the layers are all set to 1. We allow the network to include non-linearities and biases in order to simulate realistic conditions. In Figure~\ref{fig:position_weights}, we show the weights found by stochastic gradient descent training. They fit the handcrafted weights in Equation~\eqref{eq:handCraftedPositionWeights} remarkably well.

\subsection{Decoding position}

We now show that it is also possible to perform this inverse operation, in other words, starting from a position $z$, output a 1d signal which approximates a delta at position $z$. To do this, we use a triangular approximation of the Dirac delta. For a Dirac positioned at $a\in \left[ 0,\right]$, this approximation is :
\begin{equation}
y_a(t) = 
    \begin{cases}
    1-\lvert t-a\rvert, \; if \: \lvert t-a\rvert < 1\\
    0, \; \mathrm{otherwise}
    \end{cases},
\end{equation}
It is important to note that the continuous sampling of the parameter space (the position of the Dirac) and subsequent discretisation is crucial to obtaining successful decoding (as it was in the case of disks). Indeed, we also tried to use the approach described in the case of the encoder, that is to say that the Dirac is a one-hot vector at the position $a$, similar to the experiments described in the ``CoordConv'' network \cite{liu2018intriguing}. In this case, the database is limited, and the decoding is not successful. In particular, interpolating between known datapoints is quite unstable. Sampling a continuous parameter $a$ and choosing an appropriate discretisation solves this problem.

The decoding network was chosen in a similar manner to the case of disks \ref{subsec:diskDecoding}, with 1D convolutions of size $3$, biases and leaky ReLU non-linearities. The filter depths chosen were : $[1, 2, 4, 4, 4, 8]$, with an output signal size of $n=64$. The results of the decoding can be seen in Figure~\ref{fig:dirac_decoder}.

\def \dirac_decoder_width{0.8}
\begin{center}
\begin{figure*}
\begin{tabular}{cc}
~~~~~~~~~~&\includegraphics[width=\dirac_decoder_width \textwidth]{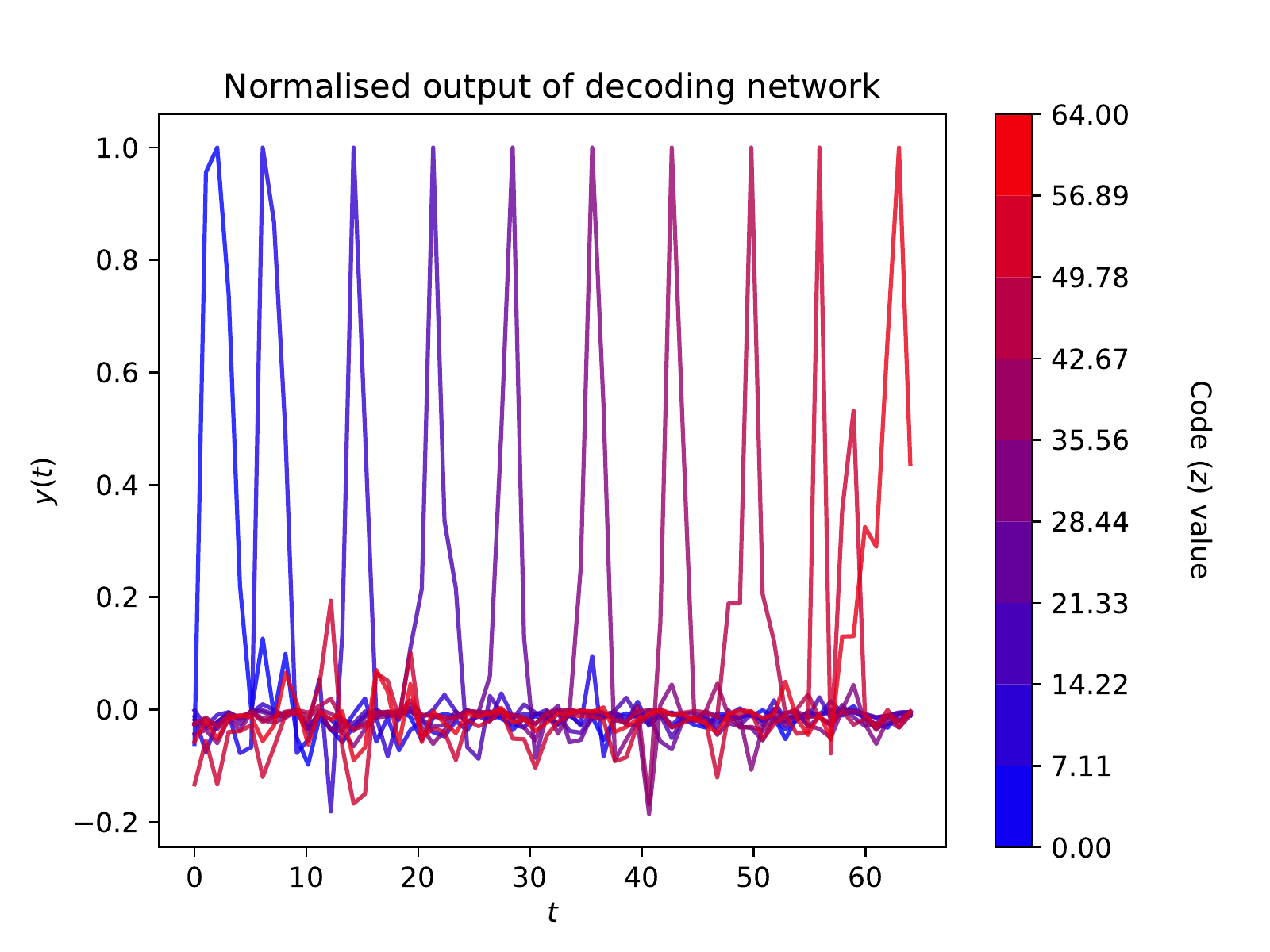}
\end{tabular}
\caption{\textbf{Normalised output of decoding of a postion to a 1D Dirac}. We show the decoding of increasing values of $z$. We have normalised each output $y$, to highlight the position of the Dirac.}
\label{fig:dirac_decoder}
\end{figure*}
\end{center}

In this Section, we have described a hand-crafted filter which, when coupled with subsampling, can achieve perfect encoding of the position of a Dirac input signal. We show that a network with an appropriate architecture indeed finds this filter during training. Secondly, we have shown experimentally that decoding is also possible as long as the latent space is sampled in a continuous manner and the corresponding signals are appropriately discretised. This highlights the necessity of correctly sampling the input data.

\section{Conclusion and future work}

We have investigated in detail the specific mechanisms which allow autoencoders to encode and decode two fundamental image properties : size and position. The first property is studied via the specific case of binary images containing disks. We first showed that the architecture we proposed was indeed capable of projecting to and from a latent space of size 1. We have shown that the encoder works by integrating over disk, and so the code $z$ represents the area of the disk. In the case where the autoencoder is trained with no bias, the decoder learns a single function which is multiplied by a scalar that is dependent on the size of the disk. Furthermore, we have shown that the optimal function is indeed learned by our network during training. This indicates that the decoder works by multiplying and thresholding this function to produce a final binary image of a disk. We have also illustrated certain limitations of the autoencoder with respect to generalisation when datapoints are missing in the training set. This is potentially problematic for higher-level applications, whose data have higher intrinsic dimensionality and therefore are more likely to include such holes. We identify a regularisation approach which is able to overcome this problem particularly well. This regularisation is asymmetrical as it consists of regularizing the encoder while leaving more freedom to the decoder.

Secondly, we have analysed how an autoencoder is able to process position in input data. We do this by studying the case of vectors containing Dirac delta functions (or ``one-hot vectors''). We identify a hand-crafted convolutional filter and prove that by using convolutions with this filter and subsampling operations, an encoding network is able to perfectly encode the position of the Dirac delta function. Furthermore, we show experimentally that this filter is indeed learned by an encoding network during training. Finally, we show that a decoding network is able to decode a scalar position and produce the desired Dirac delta function.

We believe that it is important to study generative networks in simple cases in order to properly understand how they work, so that, \emph{in fine}, we can propose architectures that are able to produce increasingly high-level and complex images in a reliable manner and with fine control over the results (for example interpolating in the latent space). An important future goal is to extend the theoretical analyses obtained to increasingly complex visual objects, in order to understand whether the same mechanisms remain in place. We have experimented with other simple geometric objects such as squares and ellipses, with similar results in an optimal code size. Another question is how the decoder works with the biases included. This requires a careful study of the different non-linearity activations as the radius increases. Finally, we are obviously interested in how these networks process other fundamental image properties, such as rotation or colour. Some recent interesting work on increasing independence in the latent codes' elements \cite{lample2017fader} could be useful in this respect.



\textbf{Acknowledgements} 
This work was funded by the Agence Nationale de la Recherche (ANR), in the MIRIAM project


\bibliographystyle{unsrt}  
\bibliography{autoencoders_arxiv}

\newpage


\begin{appendices}

\section{Creating the disk dataset}
\label{app:diskDataSet}

We wish to create a dataset which contains images of centred disks. Since the autoencoder must project each image to a continuous scalar, it makes sense to generate the disks with a continous parameter $r$, and that the disks also be ``continuous'' in some sense (each different value of $r$ should produce a different disk. For this, as we mentioned in Section~\ref{subsec:trainingDataDisk}, we create the training images $x_r$ as
\begin{equation}
x_r = g_{\sigma} \ast \indic{\mathbb{B}_r},
\end{equation}
where $\indic{\mathbb{B}_r}$ is the indicator function of the ball of radius $r$, and $g_{\sigma}$ is a Gaussian kernel with variance $\sigma$. In practical terms, we carry this out using a Monte Carlo simulation to approximate the result of the convolution of an indicator function with a disk. Indeed, let $\xi_{i, i=1 \dots N}$ be a sequence of independently and identically distributed (iid) random variables, with $\xi_i \sim \mathcal{N}\left( 0, \sigma \right)$. Each pixel at position $t$ is evaluated as
\begin{equation}
    x_r(t) = \frac{1}{N} \sum_{i=1}^{N} \indic{B_r}(\xi_i).
\end{equation}
According to the law of large numbers, this tends to the exact value of $g_{\sigma} \ast \indic{\mathbb{B}_r}$, and gives a method of producing a continuous dataset.

While other approaches are available (evaluating the convolution in the Fourier domain, for example), this is simple to implement and generalises to any shape which we can parametrise. We also note that the large majority of deep learning synthesis papers suppose that the data lie on some manifold, but this hypothesis is never checked. In our case, we explicitly sample the data in a smooth space.

\section{Decoding of a disk (network with no biases)}
\label{app:diskDecoding}

\begin{figure*}
    \begin{center}
    \includegraphics[width=0.75\linewidth]{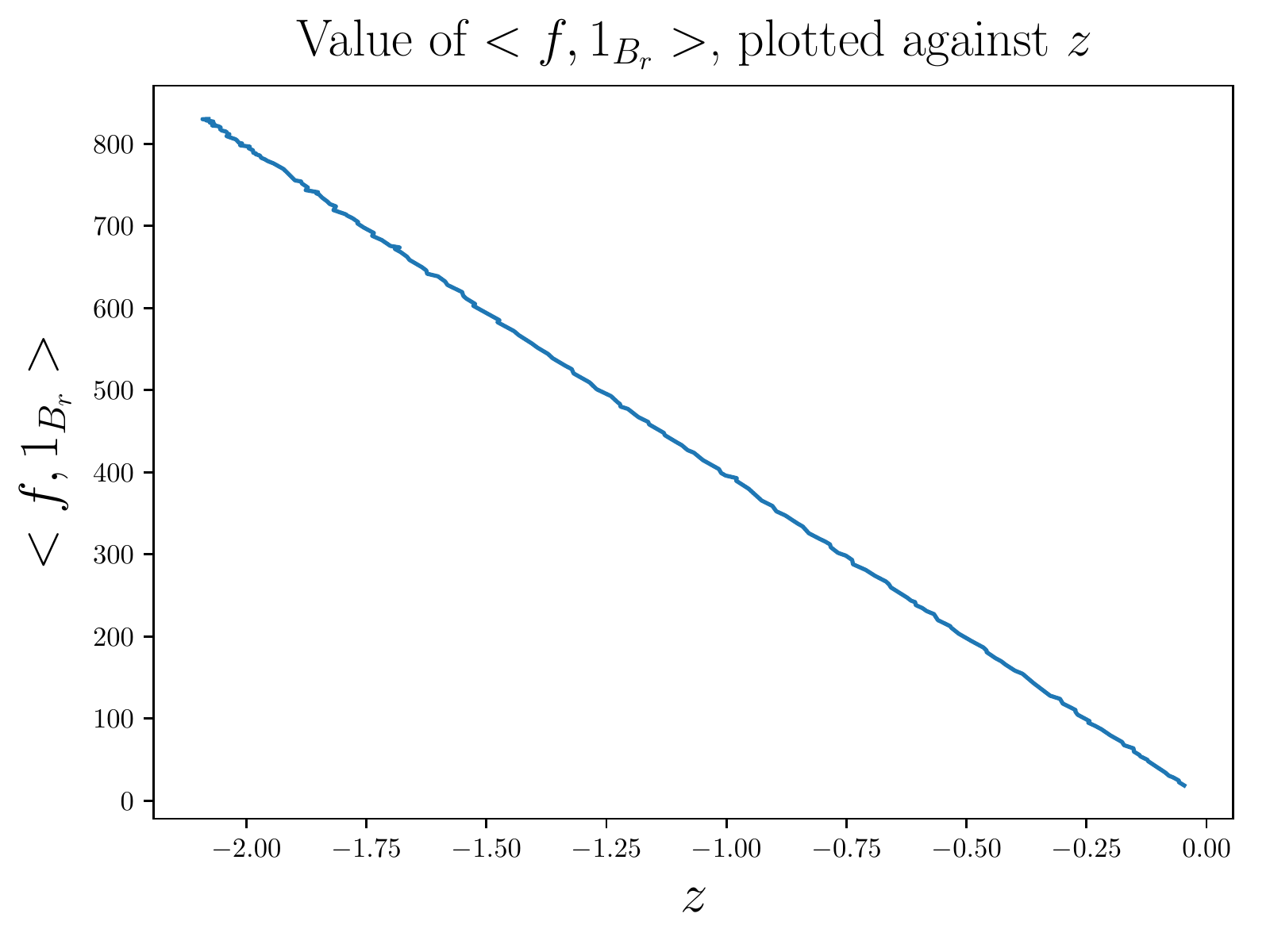}
    \end{center}

\caption{\textbf{Verification of the theoretical derivations that use the hypothesis that $y(t,r) = h(r) f(t)$ for decoding, in the case where the autoencoder contains no bias.}. We have plotted $z$ against the theoretically optimal value of $h$ ($C\left< f,\indic{B_r}\right>$, where $C$ is some constant accounting for the arbitrary normalization of $f$). This experimental sanity check confirms our theoretical derivations.}
\label{fig:noBiasHypothesisVerification}
\end{figure*}

During the training of the autoencoder for the case of disks (with no bias in the autoencoder), the objective of the decoder is to convert a scalar into the image of a disk with the $\ell_2$ distance as a metric. Given the profiles of the output of the autoencoder, we have made the hypothesis that the decoder approximates a disk of radius $r$ with a function $y(t;r) := D(E(\indic{B_r})) = h(r) f(t)$, where $f$ is a continuous function. We show that this is true experimentally in Figure~\ref{fig:noBiasHypothesisVerification} by determining $f$ experimentally by taking the average of all output profiles, and then comparing our code $z$ against its theoretically optimal value $\left< f,\indic{B_r}\right>$. We see that they are the same up to a multiplicative constant $C$.

\begin{figure*}
\begin{center}
\includegraphics[width=0.75\linewidth]{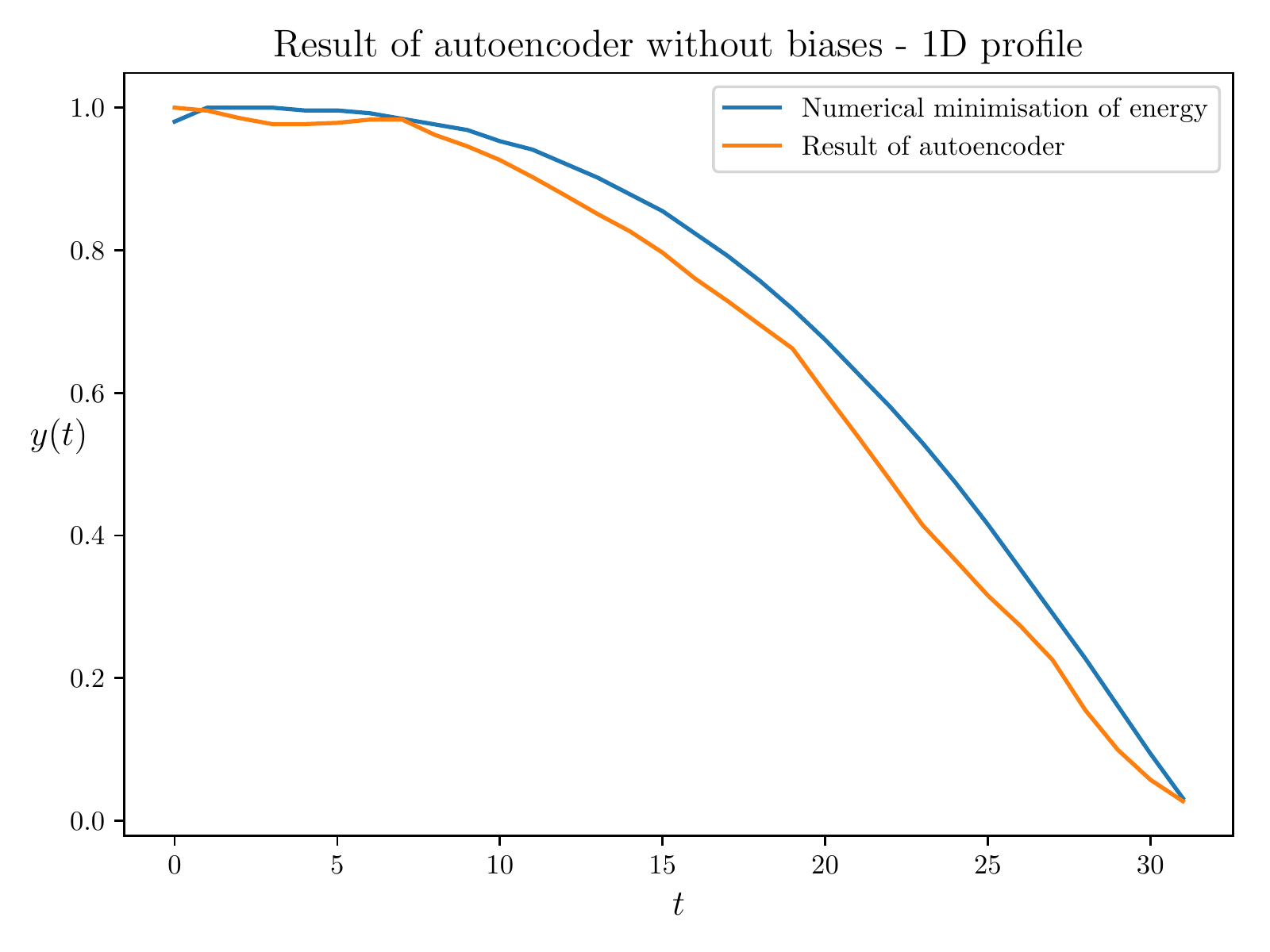}
\end{center}
\caption{\textbf{Comparison of the empirical function $f$ of the autoencoder without biases with the numerical minimisation of Equation~\eqref{eq:diskDecodingEnergyFinal}.} We have determined the empirical function $f$ of the autoencoder and compared it with the minimisation of Equation~\eqref{eq:diskDecodingEnergyFinal}. The resulting profiles are similar, showing that the autoencoder indeed succeeds in minimising this energy.}
\label{fig:energyMinimisationProfile}
\end{figure*}

We now compare the numerical optimisation of the energy in Equation \eqref{eq:diskDecodingEnergyFinal} using a gradient descent approach with the profile obtained by the autoencoder without biases. The resulting comparison can be seen in Figure~\ref{fig:energyMinimisationProfile}. One can also derive a closed form solution of Equation \eqref{eq:diskDecodingEnergyFinal} by means of the Euler-Lagrange equation and see that the optimal $f$ for Equation \eqref{eq:diskDecodingEnergyFinal} is the solution of the differential equation $y^{\prime\prime}=-kty$ with initial state $(y,y^\prime)=(1,0)$, where $k$ is a free positive constant that accommodates for the position of the first zero of $y$. This gives a closed form of the $f$ in terms of Airy functions.

\section{Autoencoding disks with a database with a limited observed radius (network with no biases)}

\label{app:diskDecodingExtrapolationFailure}

In Figure~\ref{fig:no_bias_restricted}, we see the grey-levels of the input/output of an autoencoder trained (without biases) on a restricted database, that is to say a database whose disks have a maximum radius $R$ which is smaller than the image width. We have used $R=18$ for these experiments. We see that the decoder learns a useful function $f$ which only extends to this maximum radius. Beyond this radius, another function is used corresponding to the other sign of codes (see proposition \ref{prop:lindecode}) that is not tuned.  

\def \disk_width{0.3}
\begin{figure*}
\begin{tabularx}{\textwidth}{XXX}
\includegraphics[width=\disk_width \textwidth]{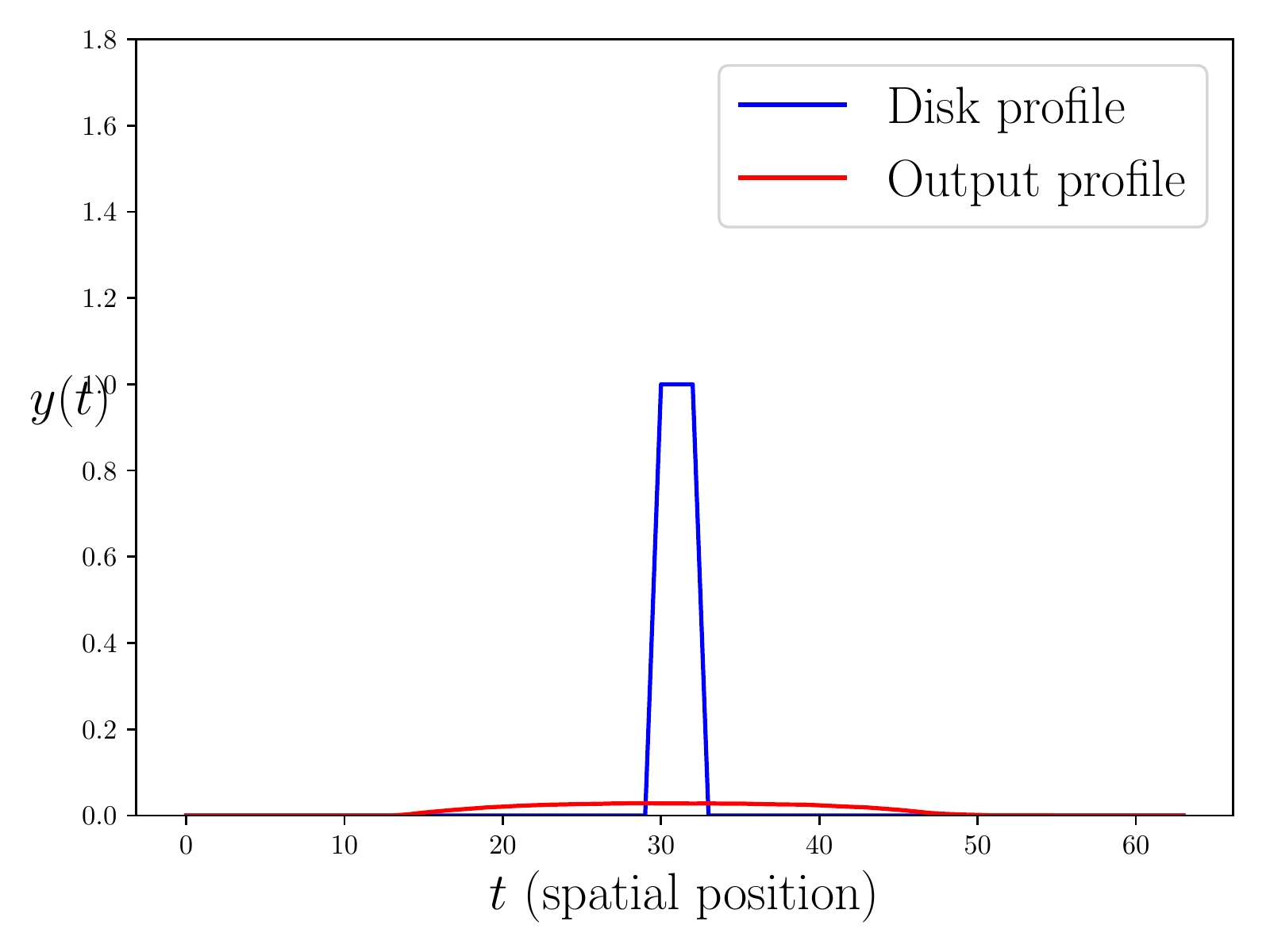}&
\includegraphics[width=\disk_width \textwidth]{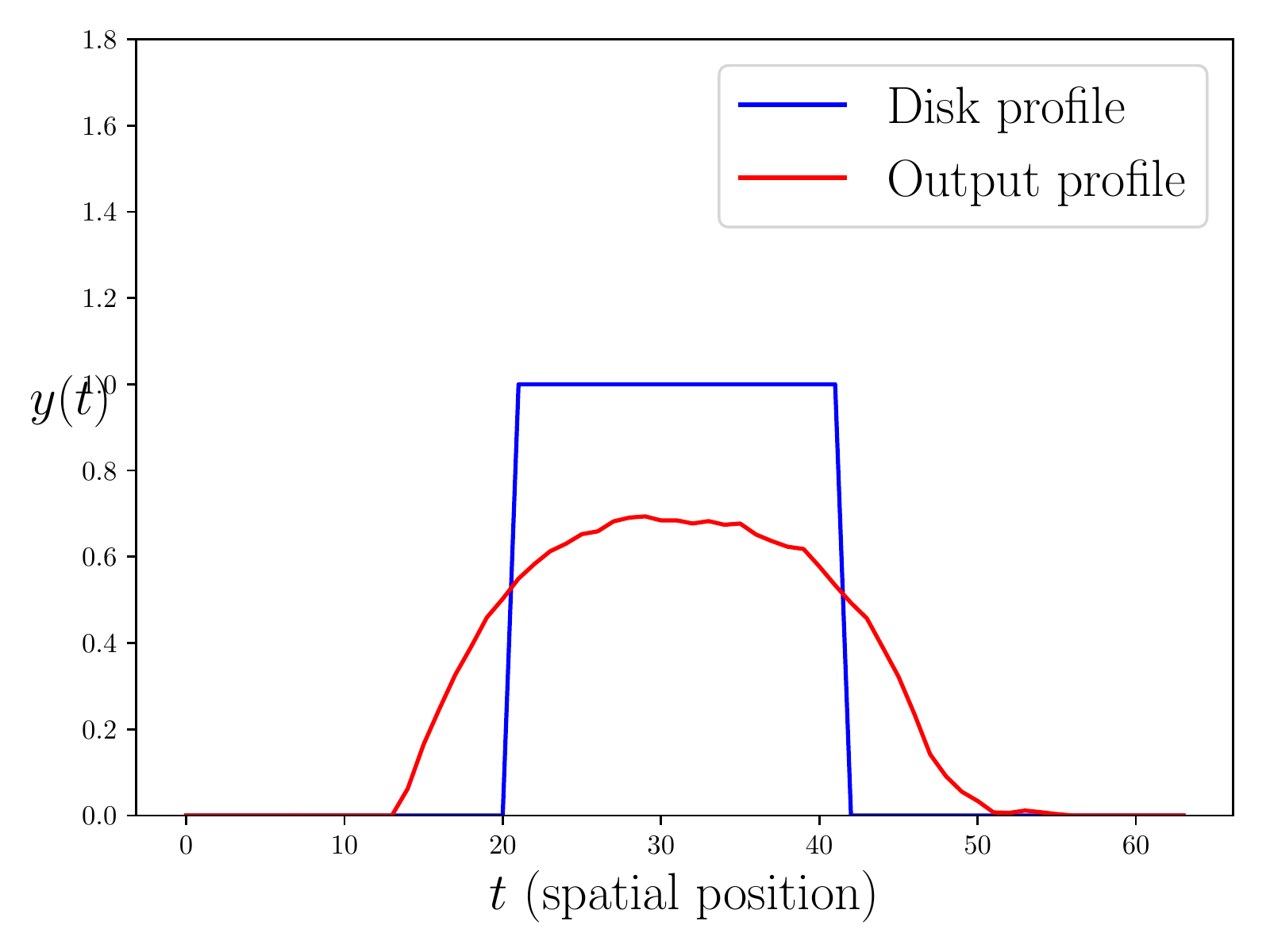}&
\includegraphics[width=\disk_width \textwidth]{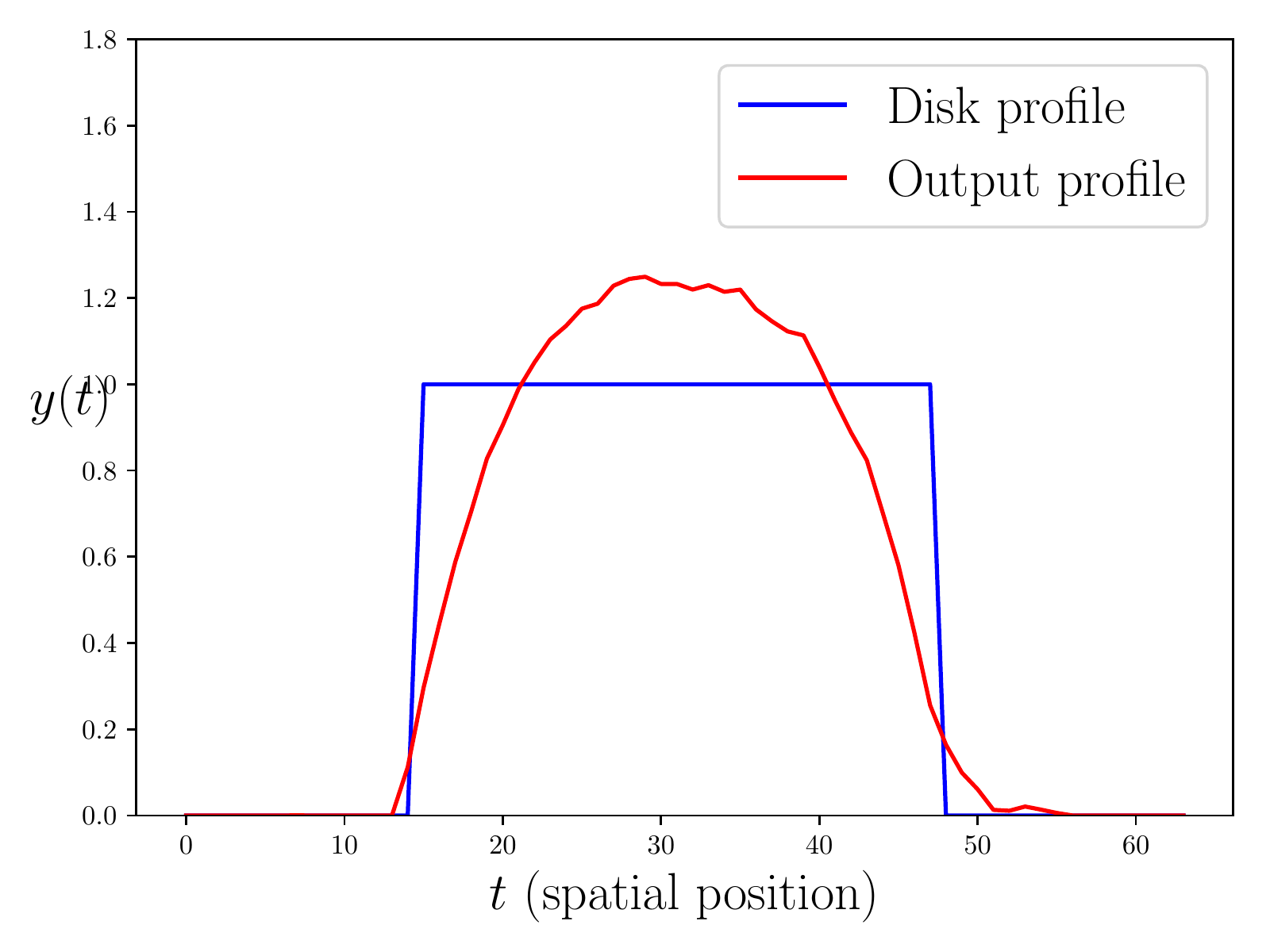}\\
\includegraphics[width=\disk_width \textwidth]{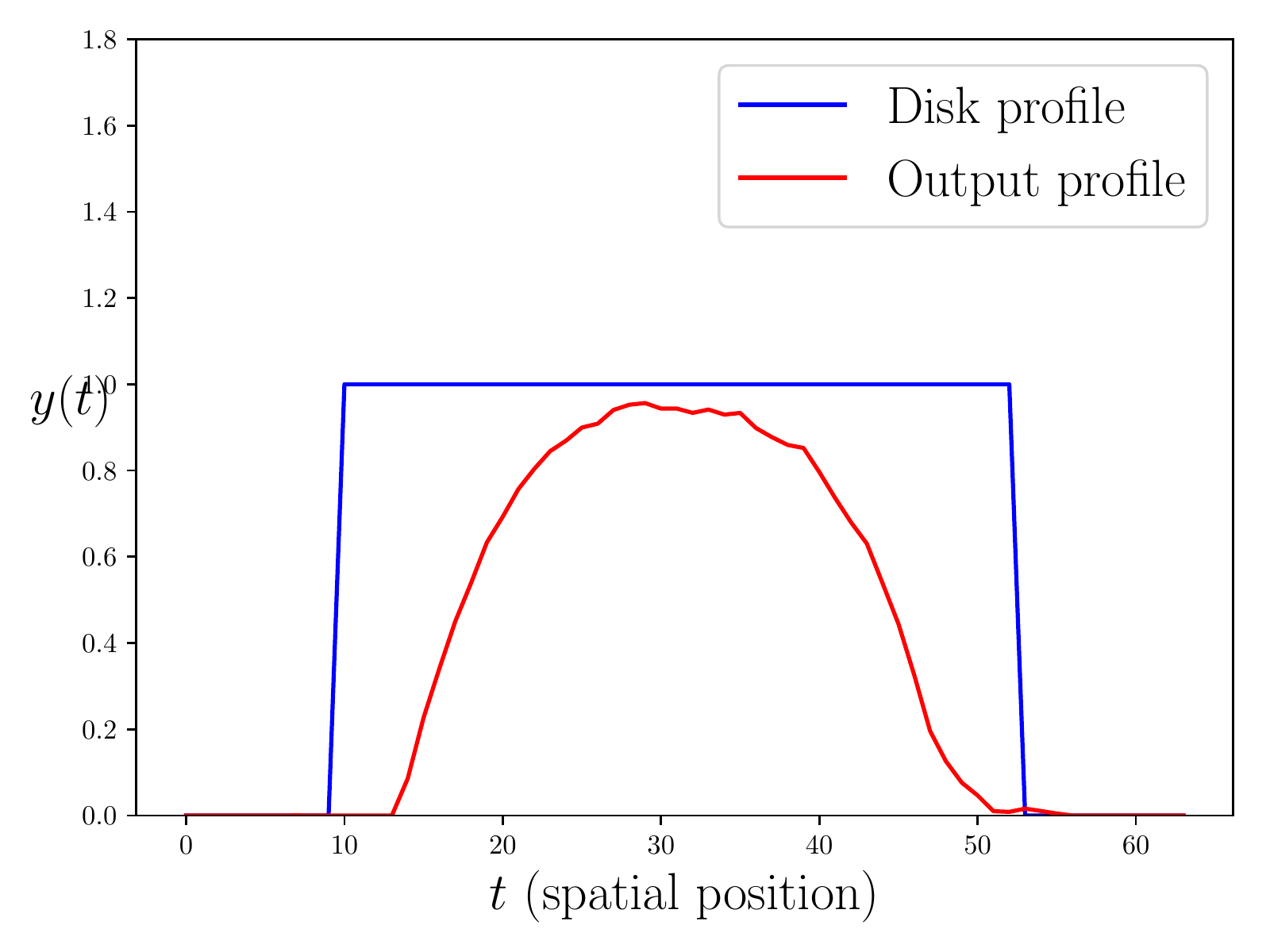}&
\includegraphics[width=\disk_width \textwidth]{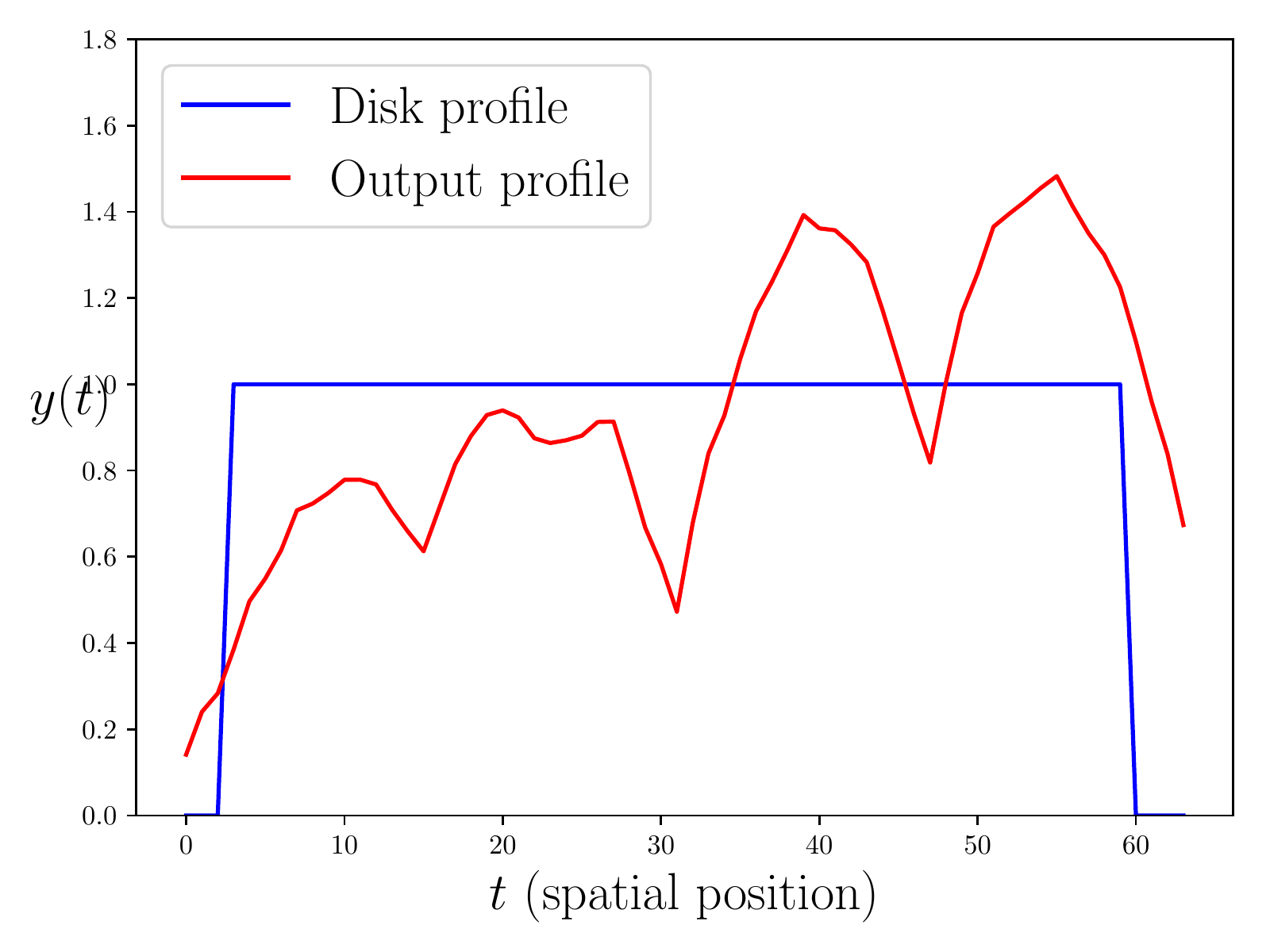}&
\includegraphics[width=\disk_width \textwidth]{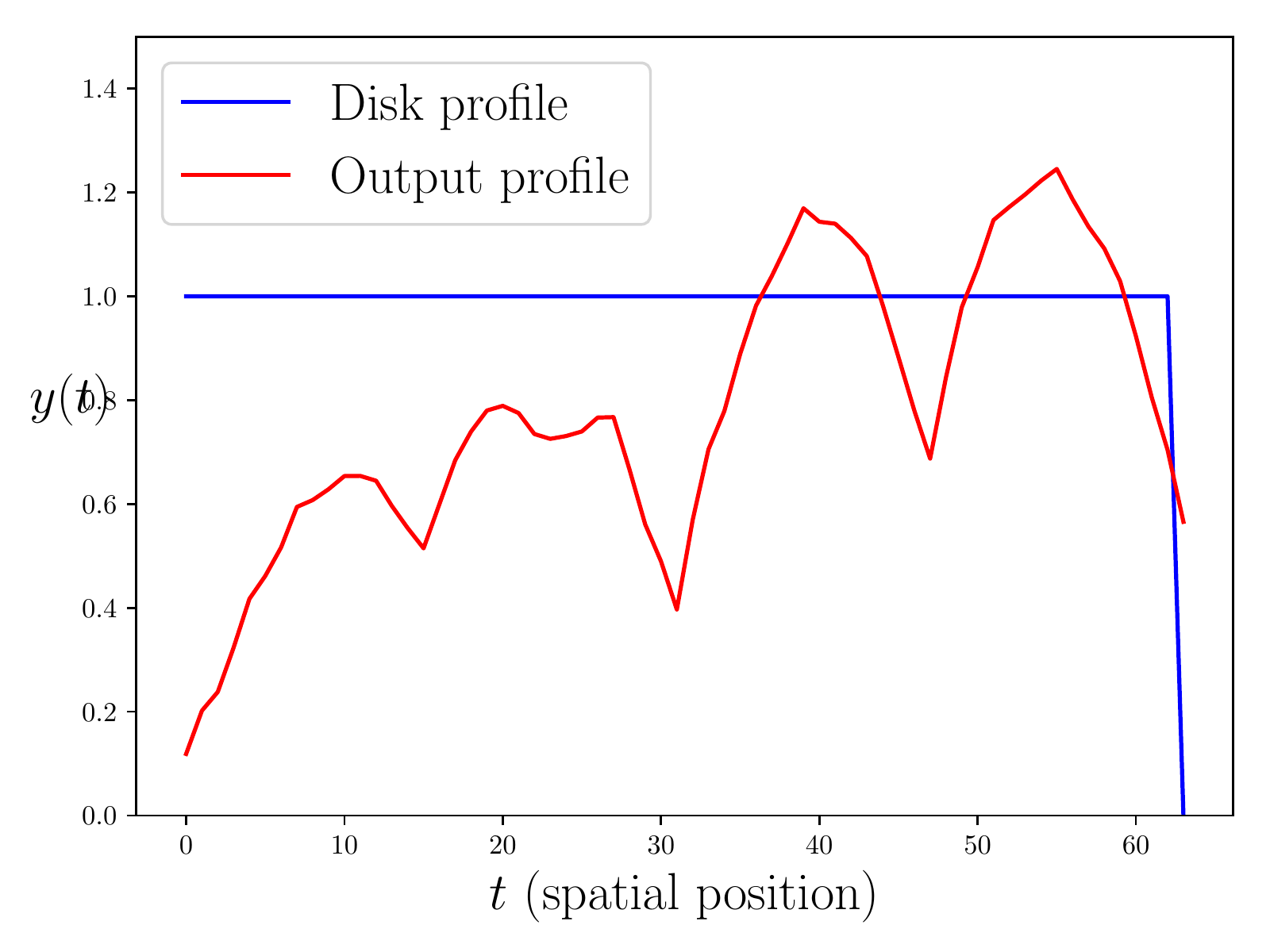}\\
\end{tabularx}
\caption{\textbf{Profile of the encoding/decoding of centred disks, with a restricted database}. The decoder learns a profile $f$ which only extends to the largest observed radius $R=18$. Beyond this radius, another profile is learned that has is obviously not tuned to any data.}
\label{fig:no_bias_restricted}
\label{fig:disk_extrapolation_no_bias}
\end{figure*}

\section{Autoencoding disks with a DCGAN \cite{radford2015unsupervised}}
\label{app:autoencodingDCGAN}

\begin{figure*}
\def \disk_radford_width{0.08}

\includegraphics[width= \textwidth]{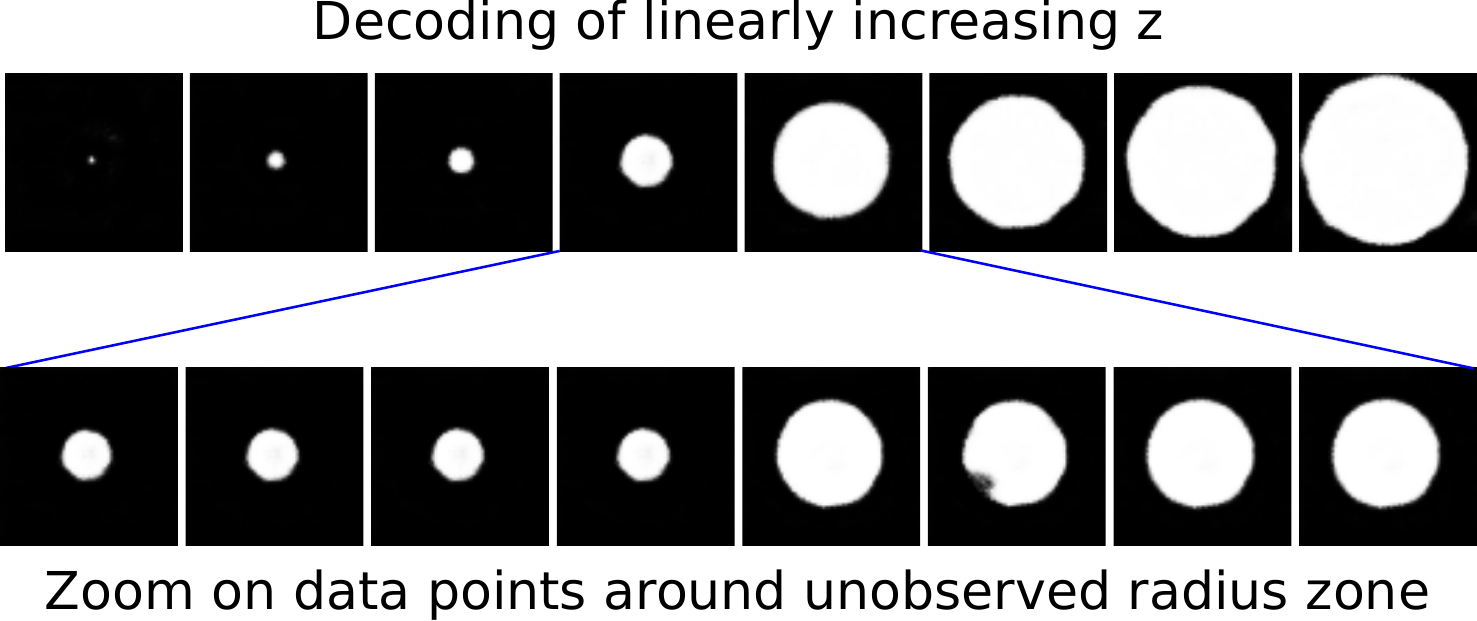}

\caption{\textbf{Output of the DCGAN of Radford et al.\cite{radford2015unsupervised} (``IGAN'') for disks when the database is missing disks of certain radii (11-18 pixels).} We can see that the DCGAN is not capable of reconstructing the disks which were not obeserved in the training dataset. This is a clear problem for generalisation. in the second we zoom on the datapoints around the radius zone which is unobserved in the training dataset.}
\label{fig:diskZhu}
\end{figure*}

In Figure~\ref{fig:diskZhu}, we show the autoencoding results of the DCGAN network of Radford et al. We trained their network with a code size of $d=1$. As can be seen, the DCGAN learns to force the training data to a predefined distribution, which cannot be modified during training (contrary to the autoencoder). Thus the network fails to correctly autoencode disks in the missing radius region which has not been observed in the training database.

\end{appendices}
\end{document}